\documentclass[10pt]{article} 
\usepackage[preprint]{tmlr}
\usepackage{comment}

\usepackage{amsmath,amsfonts,bm}

\def\eqref#1{equation~\ref{#1}}

\def\1{\bm{1}}

\DeclareMathAlphabet{\mathsfit}{\encodingdefault}{\sfdefault}{m}{sl}
\SetMathAlphabet{\mathsfit}{bold}{\encodingdefault}{\sfdefault}{bx}{n}

\usepackage{hyperref}
\usepackage{url}

\usepackage{graphicx}
\usepackage{tikz}
\usepackage{tikz-qtree}

\usetikzlibrary{trees}
\usepackage{stfloats} 
\usepackage{array}                

\usepackage{multirow}
\usepackage{booktabs}
\usepackage{float}
\usepackage{tabularray}

\usepackage{tikz}
\usepackage{tikz-qtree}
\usetikzlibrary{arrows.meta, shadows, calc}
\usepackage{booktabs, multirow, tabularx}

\usepackage{booktabs,array,colortbl,rotating,amssymb,threeparttable,multirow,graphicx,multicol}
\newcommand{\yes}{\checkmark}
\newcommand{\no}{\texttimes}

\tikzset{
  every tree node/.append style={
    draw,
    rectangle,
    rounded corners=6pt,
    minimum height=6mm,
    inner sep=2pt,
    anchor=west,
    fill=gray!5,
    drop shadow
  },
  edge from parent/.style={
    thin,
    draw=cyan!60!blue,
    -{Latex[length=2mm, width=1mm]},
    edge from parent path={
      (\tikzparentnode.east) .. controls +(.35,0) and +(-.35,0) .. (\tikzchildnode.west)
    }
  },
  every level 0 node/.append style={
    font=\bfseries\scriptsize\sffamily,
    text=black,
    minimum width=0.72in,
    text width=0.72in,
    align=center,
    fill=cyan!20
  },
  level 0/.style={
    level distance=0.8in,
    sibling distance=0.8in
  },
  every level 1 node/.append style={
    font=\bfseries\scriptsize\sffamily,
    text=blue!80!black,
    minimum width=0.7in,
    text width=0.76in,
    align=center,
    fill=blue!10
  },
  level 1/.style={
    level distance=.98in,
    sibling distance=0.18in
  },
  every level 2 node/.append style={
    font=\scriptsize\sffamily,
    text=teal!80!black,
    minimum width=1in,
    text width=1in,
    align=center,
    fill=teal!6
  },
  level 2/.style={
    level distance=1.0in,
    sibling distance=0.05in
  },
  every level 3 node/.append style={
    font=\tiny\sffamily,
    text=gray!80!black,
    minimum width=3.3in,
    text width=3.3in,
    align=center,
    fill=orange!5
  },
  level 3/.style={
    level distance=1.2in,
    sibling distance=0.01in
  }
}

\title{Memory-Augmented Transformers: A Systematic Review from Neuroscience Principles to Enhanced Model Architectures}

\author{\name Parsa Omidi \email parsa.omidi@huawei.com \\
      \addr Huawei Technologies Canada
      \AND
      \name Xingshuai Huang \email xingshuai.huang@h-partners.com \\
      \addr Huawei Technologies Canada
      \AND
      \name Axel Laborieux \email axel.laborieux@huawei.com \\
      \addr Huawei Technologies Switzerland
      \AND
      \name Bahareh Nikpour \email bahar.nikpour@h-partners.com \\
      \addr Huawei Technologies Canada
      \AND
      \name Tianyu Shi \email tianyu.shi@h-partners.com \\
      \addr Huawei Technologies Canada
      \AND
      \name Armaghan Eshaghi \email armaghan.eshaghi@huawei.com \\
      \addr Huawei Technologies Canada
      }

\def\month{MM}  
\def\year{YYYY} 
\def\openreview{\url{https://openreview.net/forum?id=XXXX}} 

\begin{document}

\maketitle

\begin{abstract}

Memory is fundamental to intelligence, enabling learning, reasoning, and adaptability across biological and artificial systems. While Transformer architectures excel at sequence modeling, they face critical limitations in long-range context retention, continual learning, and knowledge integration. This review presents a unified framework bridging neuroscience principles—dynamic multi-timescale memory, selective attention, and consolidation—with engineering advances in Memory-Augmented Transformers. We organize recent progress through three taxonomic dimensions: functional objectives (context extension, reasoning, knowledge integration, adaptation), memory representations (parameter-encoded, state-based, explicit, hybrid), and integration mechanisms (attention fusion, gated control, associative retrieval). Our analysis of core memory operations—reading, writing, forgetting, and capacity management—reveals a shift from static caches toward adaptive, test-time learning systems. We identify persistent challenges in scalability and interference, alongside emerging solutions including hierarchical buffering and surprise-gated updates. This synthesis provides a roadmap toward cognitively-inspired, lifelong-learning Transformer architectures.

\end{abstract}

\section{Introduction}
\label{intro}

Memory is fundamental to both biological and artificial intelligence (AI), serving as the foundation for cognition, reasoning, and adaptive learning \citep{camina2017neuroanatomical}. In humans, memory enables the retention, retrieval, and manipulation of information across multiple time scales, supporting complex behaviors such as decision-making and problem-solving \cite{wang2025hierarchical}. This dynamic process integrates sensory inputs, transient processing, and long-term storage, forming a sophisticated cognitive architecture.

In AI, memory has become increasingly central as models evolve from static pattern recognition to more flexible, human-like cognition. Transformer architectures \citep{ashish2017attention} have significantly advanced natural language processing, vision, and multimodal learning, yet their memory mechanisms remain restricted compared to the flexibility and efficiency of biological systems.

The primary limitation arises from self-attention’s quadratic complexity, constraining context window sizes. To stay within hardware limits, techniques like token pruning, sparse attention, and KV caching extend context but at a fidelity cost: sparse or approximate attention fractures long-range dependencies, and KV caches must evict or compress older entries, discarding vital information and harming coherence \citep{wang2024beyond}. Another issue is the static nature of knowledge representation in standard Transformers. Once trained, their parameters are fixed, lacking mechanisms for continual learning or dynamic updates. This rigidity hinders adaptation to new information or user-specific contexts and risks catastrophic forgetting when fine-tuned, unlike the flexible updating seen in biological memory.
Transformers also lag far behind biological systems in energy efficiency. The brain uses sparse, distributed, content-addressable memory with localized synaptic dynamics, operating on milliwatts of power \citep{prince2016neuromodulation, gilbert2009role}. In contrast, Transformers require intensive computation: full-context inference scales quadratically with sequence length, while autoregressive decoding must process ever-growing KV caches with linear complexity per token. This computational burden results in orders-of-magnitude higher energy consumption.
 
To bridge these gaps, memory-augmented Transformers integrate neuroscience-inspired dynamic memory mechanisms. Human memory’s efficiency and adaptability increasingly guide Transformer design, particularly its integration across timescales: sensory memory (brief stimulus retention), prefrontal cortex-maintained working memory (short-term processing), and long-term memory (lifelong learning via neocortical-hippocampal networks). This architecture balances immediate processing with stable knowledge retention, while memory allocation is further regulated by salience and context-focusing attention only on relevant inputs, as described by the global workspace theory \citep{dehaene2011global, baars2021global}.

These neuroscience-derived principles increasingly shape memory-augmented Transformer architectures. Recent models incorporate multi-timescale memory, dynamic resource allocation, and plasticity-stability trade-offs, drawing explicit inspiration from hippocampal indexing, neuromodulatory gating, and hierarchical organization.

Recent review efforts have explored memory structures in AI models from various angles. For example, \citet{ma2023memory} surveys memory augmentation techniques specifically in graph neural networks (GNNs), while \citet{du2025rethinking} offers a broader perspective, covering diverse memory mechanisms across AI models, including long-term memory, long-context memory, parametric memory modification, and multi-source memory. \citet{he2024human} approaches the topic from a human-inspired perspective, focusing on long-term memory in AI models. Other surveys narrow their scope further: \citet{shan2025cognitive} and \citet{wu2025human} focus on memory mechanisms in large language models (LLMs), while \citet{zhang2024survey} specifically investigates memory in LLM-based agents. Similarly, \citet{liu2025advances} includes a dedicated chapter on memory usage in foundation agents.

However, existing reviews are limited in two key ways. First, most rely on a single taxonomy to categorize memory-augmented methods, failing to provide a multidimensional or interdisciplinary understanding of the field. Second, many focus narrowly on specific model types or memory paradigms, such as long-term memory, LLMs, or agents, without addressing the broader landscape of memory integration across Transformer-based models.

In contrast, our review provides a comprehensive and interdisciplinary examination of memory augmentation techniques across Transformer models of various sizes, types, and applications. Our objectives are as follows:

\begin{itemize}
    \item \textbf{Establish comprehensive taxonomies} linking neuroscience principles to memory mechanisms in Transformers from three different aspects.
    \item \textbf{Analyze core memory operations}, including reading, writing, forgetting, and self-management.
    \item \textbf{Identify current challenges} in memory-augmented Transformer design and highlight emerging paradigms and future directions inspired by biological memory.
\end{itemize}

By integrating insights from neuroscience and AI, this review aims to provide a conceptual framework and practical guidance for developing more efficient, adaptive, and cognitively inspired memory-augmented Transformers.

In the following sections, we begin by introducing memory architectures in biological cognitive systems, including the structure of human memory (Section~\ref{arc_human}), interactions between different memory systems (Section~\ref{interact_memory}), and underlying computational principles (Section~\ref{comp_princ}). Section~\ref{taxonomy} presents our proposed taxonomies from three perspectives: functional objectives (Section~\ref{cat_obj}), memory types (Section~\ref{cat_type}), and integration techniques (Section~\ref{cat_int}). We then examine the mechanisms of memory operations adopted in the reviewed methods (Section~\ref{mechanism}), followed by a discussion of key challenges and future directions (Section~\ref{challenges}).

\section{Memory Architectures in Biological Cognitive Systems}
\label{memory_systems}

\begin{figure}
    \centering
    \includegraphics[width=1\linewidth]{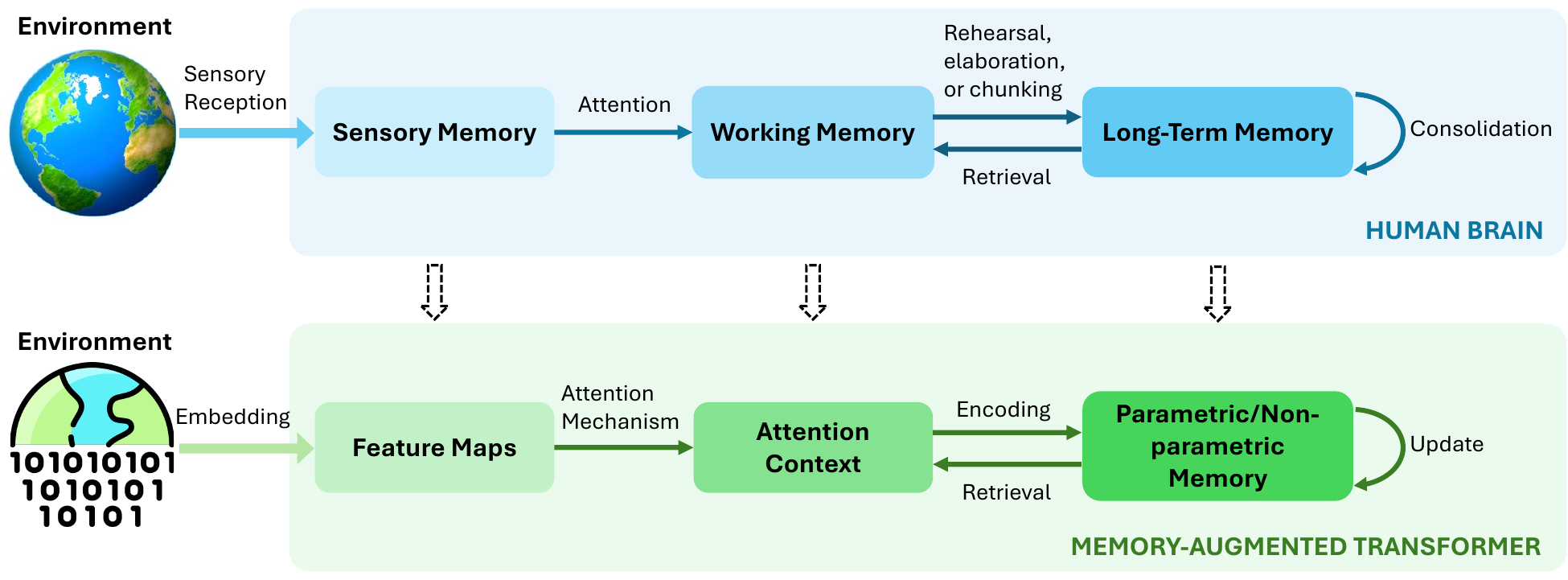}
    \caption{Parallels between the memory systems in the human brain and memory-augmented Transformers. Human memory consists of three interacting subsystems: sensory memory, working memory, and long-term memory. Memory-augmented Transformers mirror this architecture by leveraging embeddings, attention mechanisms, and advanced encoding and retrieval techniques to construct feature maps (analogous to short-term memory), attention contexts (analogous to working memory), and parametric or non-parametric memory (analogous to long-term memory).}
    \label{fig:parallel}
\end{figure}

Human memory operates as an interconnected, multi-layer network that stores, retrieves, and adapts information across several time-scales. Because these operations are hierarchical and widely distributed, stored knowledge is continuously reorganised, supporting rapid perception, flexible reasoning, and lifelong learning. This section reviews the biological architecture of memory and extracts principles that can inform cognitively-inspired AI models.

\subsection{Architecture of Human Memory}
\label{arc_human}
Rather than a single store, human memory comprises three interacting subsystems, i.e., sensory, working, and long-term memory, as shown in the upper part of Figure \ref{fig:parallel}. Each of them is optimised for a distinct combination of capacity, persistence, and processing depth \citep{cowan2008differences}. Together they enable perception, decision-making, and learning across milliseconds to decades.

\textbf{Sensory Memory: the Initial Buffer.} Sensory memory provides a high-bandwidth, ultra-short buffer for raw perceptual input: visual traces (iconic) persist for $\approx$ 250 ms and auditory traces (echoic) for up to 2–3s \citep{reznik2023dissociating}. In that brief window the brain analyses many stimuli in parallel; only items flagged by attention transition to working memory, while the rest decay rapidly, preventing overload.
Neurally, these transient traces arise from sustained activity in primary sensory cortices and thalamo-cortical loops, organised into modality-specific registers that filter noise and normalise signals before further processing \citep{camina2017neuroanatomical}.
Transformers mimic part of this stage via token embeddings and positional encodings, which stabilise raw inputs for downstream layers. Yet, unlike biological circuits that adapt gain and leverage oscillations for temporal binding, current AI pipelines remain static, making robust perception under noise and context-dependent retention an open challenge for memory-augmented models.

\textbf{Working Memory: the Cognitive Workspace.} Working memory provides a transient, capacity-limited workspace that actively maintains and manipulates information required for reasoning, problem-solving, and goal-directed behaviour \citep{miller1956magical, baddeley2003working}. Empirical estimates place its span at roughly four to seven “chunks,” a limit mitigated by chunking strategies and sustained by oscillatory activity in the prefrontal–parietal network.

Persistent firing, which is often organised through theta–gamma coupling, keeps multiple representations simultaneously accessible, while dopaminergic signals from the ventral tegmental area gate updates, suppress distractions, and prioritise task-relevant items \citep{roux2014working}. Cross-modal binding is supported by beta-band synchrony that links prefrontal cortex with hippocampal and sensory regions, enabling flexible recombination of auditory, visual, and spatial cues during complex tasks \citep{quak2015multisensory}.

Functionally, the prefrontal cortex operates as a central executive, allocating attention, switching tasks, and coordinating specialised buffers (e.g., phonological loop, visuospatial sketchpad) \cite{russin2020deep}. This distributed control balances stability with rapid updating, allowing the system to adapt to changing demands while avoiding interference.

Transformer self-attention partially echoes these operations by selectively weighting tokens within a fixed context window. Yet current models lack biologically inspired features such as neuromodulatory gating, oscillatory binding, and energy-efficient recall; external memories and recurrent variants narrow the gap but have yet to match the flexibility and robustness of human working memory.

\textbf{Long-Term Memory: the Knowledge Repository.} Long-term memory (LTM) is the brain’s durable storehouse, capable of retaining knowledge and experience for years or even a lifetime. Its defining strength is persistence: after consolidation, a trace can remain accessible indefinitely, provided it is periodically reactivated. Information is organised hierarchically into interconnected schemas that accelerate retrieval and support broad generalisation, yet the system stays plastic because each act of recall can render a trace temporarily labile and open to updating before it is re-stored during reconsolidation \citep{luo2022acquiring, lee2017update}.

Two complementary consolidation processes underpin this durability. \textbf{Synaptic consolidation}, completed within hours, strengthens hippocampal circuits through activity-dependent events such as calcium spikes and sharp-wave ripples \citep{mujawar2021memory}. \textbf{Systems consolidation} unfolds over days to years, as coordinated oscillations during sleep transfer memory indices from the hippocampus to distributed neocortical networks, creating resilient, cortex-based representations that can survive hippocampal damage \citep{luo2022acquiring}.

LTM comprises episodic and semantic subsystems. \textbf{Episodic memory} records personally experienced events tied to a specific time and place and relies on hippocampal pattern completion for cue-based recall. \textbf{Semantic memory} stores abstract facts and concepts in widely distributed cortical networks, allowing individuals to answer questions like a capital city’s name without re-living the original learning episode \citep{kumar2021semantic}. The interplay of these subsystems enables both vivid recollection and flexible inference.

Adult neurogenesis in the dentate gyrus adds further adaptability, inserting new neurons that improve pattern separation and support the incorporation of novel information without erasing older traces \citep{anacker2017adult}. This continual renewal helps the brain distinguish similar experiences and maintain cognitive flexibility across the lifespan.

Current AI systems approximate LTM with a mix of parameter-encoded knowledge and external memories. Parameter storage offers instant access but is costly to update, whereas external key–value banks such as Memformer’s fixed-size slots \citep{wu2020memformer} or EMAT’s compressed QA memories \citep{wu2022efficient} allow on-the-fly writes and reads at inference time. Retrieval-Augmented Generation (RAG) \citep{lewis2020retrieval} extends this idea by fetching fresh documents from external indices before every response, giving models a dynamic knowledge base. Despite these advances, artificial LTM still suffers from limited consolidation and vulnerability to catastrophic forgetting when new data overwrites old weights \citep{ranjith2024adaptive}. Closing this gap will require biologically inspired mechanisms, e.g., dynamic consolidation, adaptive forgetting, and hierarchical memory layouts, that mirror the robustness and context sensitivity of human long-term memory.

\subsection{Interactions Between Memory Systems}
\label{interact_memory}
Human memory functions as a dynamic network, where sensory, working, and long-term stores communicate continuously to maximize learning and behaviour. Instead of isolated modules, these systems exchange activity through converging cortical–subcortical loops that adapt to context, attention, and emotional salience.

\textbf{Encoding, Consolidation, and Retrieval.}
Encoding begins when sensory traces reach the prefrontal cortex, which filters and amplifies task-relevant inputs before they flow into working and long-term stores. Subsequent consolidation—particularly during slow-wave sleep—relies on hippocampal replay that drives neocortical reorganisation, stabilising both episodic and semantic traces \citep{klinzing2019mechanisms}. Retrieval completes this cycle: a partial cue reactivates hippocampal indices, triggering pattern-completion processes that reconstruct the distributed cortical representation and return it to working memory for use or further updating \cite{teyler2007hippocampal}.

\textbf{Top-Down and Bottom-Up Modulation.}
During retrieval, top-down signals from prefrontal regions bias processing toward current goals, suppressing irrelevant information, while bottom-up inputs from sensory and limbic areas flag novelty or emotional significance. Neuromodulators such as dopamine and acetylcholine strengthen synapses that encode behaviourally important events, fine-tuning what is stored or updated \citep{gazzaley2012top}.

\textbf{Emotional and Multimodal Integration.}
Emotionally charged or multisensory experiences recruit coordinated activity in the amygdala, hippocampus, and prefrontal cortex, yielding more persistent memories \citep{dolcos2004interaction}. The thalamus binds inputs from different senses, and hippocampal pattern completion links them into context-rich episodes that can be triggered later by a single cue.

\textbf{Competitive and Co-operative Dynamics.}
Memory systems shift between competition and cooperation. Under stress or heavy cognitive load, control can pass from flexible hippocampal networks to faster, habit-based striatal circuits, ensuring rapid action \citep{schwabe2013stress}. In calmer conditions, episodic and semantic stores collaborate: detailed recollections supply context while abstract schemas guide generalisation and planning \citep{moscovitch2016episodic}.

\textbf{Default Mode Network and Predictive Processing.}
The Default Mode Network supports offline consolidation, autobiographical recall, and mental simulation \citep{higgins2021replay}. By replaying prior experiences, it updates internal models, enabling predictive processing that helps the brain (and, by extension, AI agents) anticipate future events and adapt behaviour accordingly \citep{liu2021experience}.

Understanding this balance of stability and plasticity offers a template for AI: memory architectures that coordinate fast buffers with slower, more permanent stores, employ selective gating, and integrate cross-modal information can move beyond static storage toward lifelong, context-aware learning.

\subsection{Computational Principles from Biological Memory}
\label{comp_princ}
The architectural and functional properties of biological memory systems reveal fundamental computational principles that guide memory-augmented transformer design. These principles address universal computational challenges: managing limited resources, balancing stability with plasticity, and coordinating information flow across multiple timescales. Abstracting these neurobiological solutions into engineering heuristics yields a practical design playbook, now guiding the development of the most effective memory-augmented Transformer architectures.

\textbf{Hierarchical Resource Allocation.} Biological memory demonstrates that computational efficiency emerges from hierarchical organization rather than uniform processing \citep{hasson2015hierarchical}. Sensory memory's high-bandwidth, ultra-short retention enables parallel pre-processing, working memory's capacity-limited workspace allows flexible manipulation, and long-term memory's distributed storage supports both rapid recall and gradual consolidation. Multimodal evidence suggests that these hierarchical dynamics emerge as a global organizing principle of mammalian brains, with cortical timescale gradients topographically mirrored in striatum, thalamus, and cerebellum \citep{raut2020hierarchical}. \textit{This hierarchical structure suggests that artificial systems benefit from multi-tier memory architectures that match storage characteristics to computational demands}.

\textbf{Attention-Memory Bidirectional Coupling.} The interaction between attention and memory reveals a crucial computational principle: memory systems both shape and are shaped by attentional mechanisms \citep{chun2007interactions}. Extensive evidence demonstrates that attention and memory cannot operate without each other: memory has limited capacity and attention determines what will be encoded, while memory from past experience guides what should be attended. Brain areas important for memory, such as the hippocampus and medial temporal lobe structures, are recruited in attention tasks, and memory directly affects frontal-parietal networks involved in spatial orienting. This bidirectional coupling enables adaptive resource allocation and context-sensitive processing through attention-dependent coupling between forebrain and brainstem neuromodulatory systems \citep{cicero2025attention}. \textit{These principles suggest AI memory systems should incorporate feedback loops between retrieval mechanisms and encoding processes}.

\textbf{Neuromodulatory Gating and Significance Filtering.} Biological memory formation relies on the interplay between Hebbian plasticity and neuromodulatory systems, making memory encoding inherently state-dependent and gated by the behavioral significance of information \citep{bazzari2019neuromodulators}. Neuromodulators such as dopamine and acetylcholine have distinct and complementary roles: dopamine regulates the induction of synaptic plasticity by modulating glutamatergic signaling, while acetylcholine orchestrates neuronal activity at both synaptic and network-wide levels. These systems establish computational principles of selective attention and adaptive thresholding, which allow the brain to prioritize salient information for encoding. Notably, selective neuromodulatory gating reflects a fundamental asymmetry in biological cognition: less than 5\% of brain activity is devoted to conscious processes, while over 95\% operates unconsciously, thus maximizing efficiency and resource allocation \cite{raichle2001default} \cite{nail2021most}. The strict limitations of working memory—estimated at 4–7 meaningful chunks—essentially define the conscious mind's computational budget. \textit{This insight suggests that artificial memory systems should carefully reserve costly, conscious-like processing for high-priority tasks such as novelty detection or conflict resolution, while routine memory operations are best delegated to automatic, parallel processing pathways analogous to the brain’s unconscious majority}.

\textbf{Replay-Based Consolidation and Interference Management.} The brain's solution to the stability-plasticity dilemma through dual-phase consolidation, i.e., rapid hippocampal encoding followed by gradual neocortical integration, reveals essential computational principles for managing memory interference \citep{squire2015memory}. Neural replay during sleep drives consolidation by reactivating patterns of network activity that occurred during previous experience, leading to potentiation of relevant synaptic connections in the cortex. This process enables rapid learning without catastrophic forgetting through replay-based consolidation and systems-level reorganization, where hippocampal replay propagates to cortex with reprocessing to extract statistical overlap from different encoding episodes. The stability-plasticity dilemma reflects a fundamental challenge in learning systems: retaining stored memory while learning new information \citep{mermillod2013stability}. \textit{The implication for AI is that effective memory systems require complementary fast and slow learning mechanisms with explicit consolidation phases}.

\textbf{Content-Addressable Associative Retrieval.} Biological memory systems excel at content-addressable retrieval through associative networks that enable pattern completion from partial cues \citep{rolls2013mechanisms}. The hippocampal CA3 subfield functions as an autoassociative network that stores experiences as memories, with abundant recurrent connections exhibiting spike-timing-dependent plasticity that allows pattern completion and recovery of stored patterns from noisy cues \citep{kang2024distinguishing}. Empirical evidence from direct hippocampal recordings reveals pattern completion mechanisms where reinstatement of encoding patterns occurs during successful recollection, linked to gamma power fluctuations that coordinate selection of target-relevant neurons \citep{staresina2016hippocampal}. The CA3 region exemplifies this through its ability to retrieve complete episodic memories from fragmentary input, while semantic memory supports flexible inference through conceptual associations mediated by distributed cortical networks. \textit{These mechanisms suggest that artificial memory architectures should prioritize associative rather than positional indexing and support similarity-based retrieval that mirrors biological pattern completion processes}.

\textbf{Cross-Modal Integration and Binding.} Biological memory systems demonstrate sophisticated cross-modal integration capabilities essential for unified cognitive processing \citep{nyhus2010functional, shi2023neural}. Theta and gamma oscillations enable interaction between cortical structures and the hippocampus for encoding and retrieval of episodic memories, where cortical gamma oscillations bind relevant stimulus features for perceptual representations and gamma phase synchronization between cortical and hippocampal neurons provides the mechanism for encoding diverse cortical information into hippocampal representations \citep{nyhus2010functional}. Evidence from spatial decision-making tasks reveals that hippocampal-prefrontal interactions show maximum coherence during cross-modal binding, with theta rhythm dynamically modulating neurons in both regions \citep{tavares2022hippocampal}. Cross-modal prediction is supported by indirect pathways mediated by higher-order areas that receive convergent sensory inputs \citep{shi2023neural}, \textit{suggesting that artificial memory systems should incorporate mechanisms for binding information across modalities through oscillatory coordination}.

These computational principles collectively point toward memory-centered cognitive architectures where memory systems serve as the substrate for all cognitive operations rather than passive storage devices. The memory-centered cognition perspective places an active association substrate at the heart of cognition, making prediction and priming based on prior experience fundamental aspects of processing. Understanding these principles provides the foundation for designing artificial memory systems that move beyond static storage toward dynamic, adaptive, and context-aware memory architectures that can support the flexible, hierarchical, and associative processing characteristic of biological cognition.

\section{Taxonomy of Memory-Augmented Transformers}
\label{taxonomy}
Memory-augmented Transformers aim to overcome the fixed-context and static-knowledge constraints of standard models by drawing inspiration from the dynamic nature of human memory. This section presents a taxonomy of existing architectures along three dimensions: functional objectives, memory types, and integration technique (Figure~\ref{fig:taxonomy}). We relate these categories to biological memory principles to show how they help bridge the gap between current Transformers and human-like cognition.


\begingroup
\hypersetup{hidelinks} 

\begin{figure*}[!t]
  \centering
  \makebox[\textwidth][l]{%
    \begin{tikzpicture}[grow'=right]
      \Tree
        [. {Taxonomy of Memory-Augmented Transformers}
           [.{Categorization by Functional Objectives}
             [.{Temporal Context Extension} 
             [.{Transformer-X \citep{dai2019transformer}, Compressive Transformer~\citep{rae2019compressive}, Memformer~\citep{wu2020memformer}, LongMem~\citep{wang2023augmenting}, MemLong \citep{liu2024memlong}, EM-LLM \citep{fountas2024human}, ARMT \citep{rodkin2024associative}, MemWalker \citep{chen2023walking}, Memorybank \citep{zhong2024memorybank}, MemoryLLM \cite{wang2024memoryllm},  M+ \cite{wang2025m+}, $R^{3}$mem \cite{wang2025r}} ] 
             ]
             [.{Out-of-Distribution Learning \& Adaptation} 
                [.{NAMMs \citep{cetin2024evolved}, RA-DT \citep{schmied2024retrieval}, Transformer-Squared \citep{sun2025transformer}, AdaTape \citep{xue2023adaptive}, Titans \citep{behrouz2024titans}, ATLAS \citep{behrouz2025atlas}, zip2zip \cite{geng2025zip2zip} } ]
             ]
             [.{Reasoning Enhancement}
                [.{MemReasoner \citep{ko2024memreasoner}, MATTER \citep{lee2024matter}, Memorizing Transformer \citep{wu2022memorizing}, ARMT \citep{rodkin2024associative}, LM2 \citep{kang2025lm2}, SAM \citep{le2020self}, ATLAS \citep{behrouz2025atlas}} ]
             ]
             [.{Knowledge Integration}
                [.{EMAT \citep{wu2022efficient}, MATTER \citep{lee2024matter}, DSI \citep{tay2022transformer}, Memory$^{3}$ \citep{yang2024memory3}, HippoRAG \citep{gutierrez2024hipporag}, RETRO \citep{borgeaud2022improving}, CDMem \citep{gao2025efficient} , MemoryOS \cite{kang2025memory}} ]
             ]
             [.{Task-Specific Skill Acquisition}
                [.{MeMOTR \citep{gao2023memotr}, MALT Diffusion \citep{yu2025malt}, MemBART \citep{wu2022stateful} } ]
             ]
           ]
           [.{Categorization by memory types}
             [.{Parameter-Encoded} 
                [.{DSI \citep{tay2022transformer}, Schrödinger’s Memory \citep{wang2024schrodinger}, Transformer-Squared \citep{sun2025transformer}, Titans \citep{behrouz2024titans}, ATLAS \citep{behrouz2025atlas}, Peripheral Memory \cite{zhaiperipheral} } ]
             ]
             [.{State-Based}
                [.{Transformer-XL \citep{dai2019transformer}, Compressive Transformer~\citep{rae2019compressive}, HMT \citep{he2024hmt}, TransformerFAM \citep{hwang2024transformerfam}, WorkMATe \citep{kruijne2021flexible}, MemBART \citep{wu2022stateful}, Memformer~\citep{wu2020memformer}, zip2zip \cite{geng2025zip2zip}, RMoE \citep{qiu2024layerwise} } ]
             ]
             [.{Explicit Storage} 
                [.{Memformer~\citep{wu2020memformer}, EMAT \citep{wu2022efficient}, MATTER \citep{lee2024matter}, MemGPT \citep{packer2023memgpt}, CDMem \citep{gao2025efficient}, A-MEM \citep{xu2025mem}, Memory Layers at Scale \citep{berges2024memory}, Mem0 \citep{chhikara2025mem0}, Think-in-Memory \citep{liu2023think}, MemLLM \citep{modarressi2024memllm}, MemLong \citep{liu2024memlong}, MemoryLLM \cite{wang2024memoryllm},  M+ \cite{wang2025m+}} ]
             ]
             [.{Hybrid and Multi-scale}
                [.{LM2 \citep{kang2025lm2}, Titans \citep{behrouz2024titans}, MATTER \citep{lee2024matter}, ATLAS \citep{behrouz2025atlas}, NAMMs \citep{cetin2024evolved}, MemGPT \citep{packer2023memgpt} } ]
             ]
           ]
           [.{Categorization by Integration Techniques}
             [.{Attention-Based Fusion}
                [.{Memformer~\citep{wu2020memformer}, EMAT \citep{wu2022efficient}, TransformerFAM \citep{hwang2024transformerfam}, LongMem~\citep{wang2023augmenting}, MATTER \citep{lee2024matter}, Memorizing Transformer \citep{wu2022memorizing} } ]
             ]
             [.{Gated Control Mechanisms} 
                [.{Titans \citep{behrouz2024titans}, RA-DT \citep{schmied2024retrieval}, MeMOTR \citep{gao2023memotr}, NAMMs \citep{cetin2024evolved}, RMoE \citep{qiu2024layerwise} } ]
             ]
             [.{Associative Memory Integration} 
                [.{ARMT \citep{rodkin2024associative}, AiT \citep{sun2023associative}, MemReasoner \citep{ko2024memreasoner} } ]
             ]
           ]
        ]
    \end{tikzpicture}%
  }
  \caption{Taxonomy of Memory-Augmented Transformers.}
  \label{fig:taxonomy}
\end{figure*}
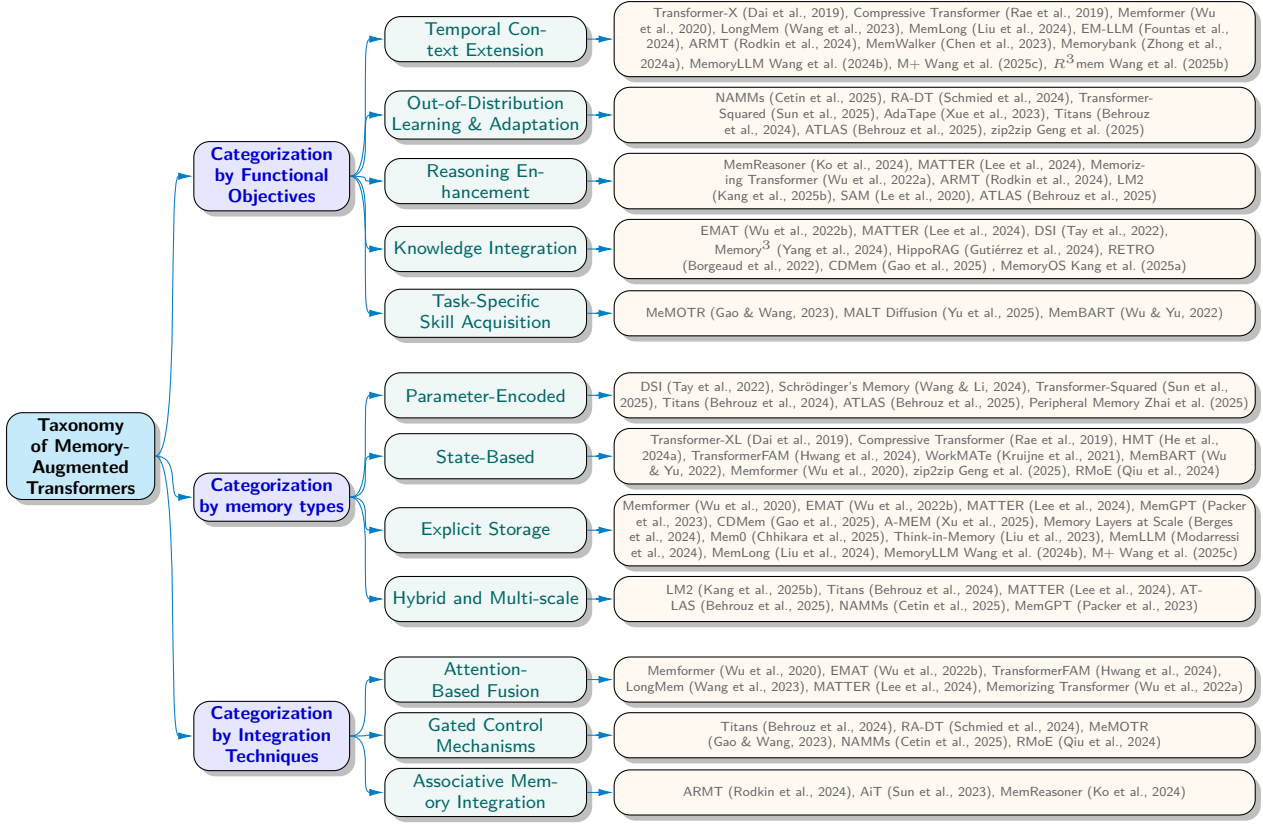
\endgroup

\subsection{Categorization by Functional Objectives}
\label{cat_obj}
Memory-augmented Transformers address fundamental AI challenges, each mapped to a distinct functional objective:

\textbf{Temporal Context Extension}. Transformers struggle to process sequences beyond a fixed window, unlike the human brain's ability to integrate experiences over long timescales. The evolution of temporal context extension reveals a clear trajectory from static windowing mechanisms toward sophisticated, biologically-inspired adaptive memory systems. 

Sliding Window Attention (SWA) \citep{beltagy2020longformer} establishes the foundational approach to linear-complexity context extension, where each token attends to a fixed window of \( w \) neighboring tokens, achieving \( \mathcal{O}(n.w) \) complexity while maintaining parallelization. However, SWA operates as a static sensory buffer without adaptive selection or contextual awareness, limiting its effectiveness for complex temporal dependencies.

The field has witnessed a clear evolutionary trajectory from this static windowing toward adaptive, memory-augmented mechanisms that progressively incorporate working memory principles. ABC (Attention with Bounded-Memory Control) \citep{peng2021abc} transforms static windowing through learned, contextualized control strategies that dynamically determine token retention within fixed memory budgets. TransformerFAM \citep{hwang2024transformerfam} introduces feedback attention loops creating sustained activations across unlimited contexts, effectively transforming static windows into dynamic working memory systems. NAMMs \citep{cetin2024evolved} employ STFT spectrogram analysis with genetic algorithms to evolve attention patterns for zero-shot cross-modal transfer, while AdaTape \citep{xue2023adaptive} extends this through adaptive tape tokens that dynamically adjust sequence content and computational allocation. ATLAS \citep{behrouz2025atlas} represents the culmination of this trend, adding sophisticated memory mechanisms after each sliding window through the Omega rule and polynomial feature mapping, achieving super-linear memory capacity.

Building upon these adaptive windowing foundations, practical implementations began with simple KV caching mechanisms. Transformer-XL \citep{dai2019transformer} introduced the concept of caching key-value pairs from previous segments with relative positional encoding, where the cache itself functions as the memory, storing compressed representations of past context to extend processing beyond fixed windows. This established the fundamental principle that memory in transformers is essentially intelligent caching. Successive approaches focused on making this cache more selective and efficient at storing important information. Compressive Transformer \citep{rae2019compressive} enhanced the basic KV cache through vector compression, boosting temporal range by 38\% by intelligently compressing older cached states rather than simply discarding them. However, it still operated under fixed storage constraints, requiring memory eviction strategies. MemoryLLM \cite{wang2024memoryllm} elevated the idea by adding a learnable write-gate, compression‐on-evict, and a neural router that selects the top-k relevant keys, enabling $\approx$ 20 k-token context with near-constant compute. M+ \cite{wang2025m+} then removes this 20 k ceiling: it splits the cache into a small on-GPU working store and a large CPU-resident long-term bank, coordinated by a co-trained retriever and read-write scheduler. The hierarchy preserves the compress-on-evict principle yet sustains coherent generation across >160 k tokens while adding <3\% throughput overhead. 
$R^{3}$mem \cite{wang2025r} introduces a reversible compression architecture that enables bidirectional transformation between raw context and compressed memory representations through hierarchical chunking across multiple semantic levels—segmenting content from paragraphs to sentences to sub-sentence units—ensuring both efficient compression and faithful reconstruction of long contexts while maintaining semantic coherence across compression-decompression cycles.

Memformer \citep{wu2020memformer} represented a breakthrough by decoupling computation from memory through similarity-based cache management—the cache became truly adaptive, updating based on content relevance rather than simple temporal recency. Its MRBP optimization cut training memory costs by 55\% by learning which cached representations were most valuable to retain.
LongMem \citep{wang2023augmenting}  and MemLong \citep{liu2024memlong} further refined cache intelligence: LongMem freezes the backbone LLM and uses a trainable SideNet to selectively retrieve and fuse the most relevant cached key-value pairs from a growing memory bank, while MemLong employs Retrieval Causal Attention to actively prune less important cached entries, demonstrating that the cache can learn what to forget as well as what to remember.

EM-LLM \citep{fountas2024human} represents the pinnacle of intelligent caching through episodic memory segmentation. It uses Bayesian surprise detection to partition the cache into meaningful episodes, enabling retrieval that combines both semantic similarity and temporal contiguity across sequences up to 10 million tokens. ARMT \citep{rodkin2024associative} scales this concept to 50 million tokens through Hopfield-inspired associative caching with explicit erase operations, while MemWalker \citep{chen2023walking} creates hierarchical cache structures using trees of text summaries, and Memorybank \citep{zhong2024memorybank} implements cognitively-inspired cache decay following human memory patterns like the spacing effect.

This progression mirrors the hierarchical integration of biological memory systems, where simple sensory buffering (SWA, basic KV caching) evolves into sophisticated working memory mechanisms (adaptive windowing, intelligent cache management) that balance immediate processing needs with longer-term contextual understanding, moving toward truly cognitive attention systems that integrate multiple timescales of temporal context.

\textbf{Out-of-Distribution (OOD) Learning and Adaptation}. Memory-augmented Transformers address the challenge of adapting to novel data distributions while preserving performance on familiar content through surprise-driven mechanisms that mirror biological memory systems' ability to encode novel experiences while maintaining stable knowledge representations \citep{barry2024fast, sinclair2021prediction, frank2022experiencing}. 

Surprise-driven adaptation forms the core of effective OOD learning. EM-LLM \citep{fountas2024human} demonstrates this through Bayesian surprise detection and graph-theoretic boundary refinement to segment sequences into episodic events. This training-free approach automatically detects distribution shifts, creating distinct memory episodes for novel patterns while preserving performance on familiar content, embodying how novelty detection enables rapid adaptation to unexpected domains without compromising existing knowledge.

Titans \citep{behrouz2024titans} advance surprise-driven adaptation through prediction error gating, where KL divergence thresholds at the single-token level determine when memory updates occur. This token-by-token surprise detection enables fine-grained, test-time learning without parameter modification, allowing models to selectively memorize novel information while avoiding interference with established knowledge. ATLAS \citep{behrouz2025atlas} similarly employs surprise signals through the Omega rule at the local context level, using sliding windows to determine which multi-token contexts warrant long-term memorization based on prediction error magnitudes across sliding windows rather than individual tokens. Dynamic Input Pruning \citep{federici2024efficient} achieves zero-parameter adaptation through magnitude-based pruning of MLP activations per token, implementing a predictor-free strategy for real-time efficiency improvements without model retraining.

zip2zip \cite{geng2025zip2zip} demonstrates adaptive tokenization as a novel OOD adaptation strategy, dynamically expanding vocabulary at inference time through compression-based token merging that enables models to efficiently process unfamiliar token patterns and domains without retraining, achieving significant latency improvements while maintaining adaptability to new linguistic distributions.

Evolutionary and adaptive mechanisms enable cross-domain generalization without domain-specific training. NAMMs \citep{cetin2024evolved} demonstrate evolutionary optimization of attention patterns using genetic algorithms and STFT spectrogram analysis to evolve token retention policies for zero-shot cross-modal transfer. AdaTape \citep{xue2023adaptive} extends adaptive allocation through elastic input sequences with adaptive tape tokens that dynamically adjust sequence content and computational allocation based on problem complexity. RA-DT \citep{schmied2024retrieval} combines episodic memory with surprise-based pruning to retain high-error experiences, boosting multitask efficiency by 40\% while mimicking dopaminergic learning mechanisms. Transformer-Squared \citep{sun2025transformer} encodes procedural expertise directly into parameter space using SVD, dynamically blending expert vectors during inference to reach 90\% accuracy on unseen tasks, despite a 15\% latency overhead. \citet{dutta2024memory} extends the Memformer architecture to procedural computation by storing and combining past optimization gradients with learned coefficients, achieving 98\% convergence accuracy on OOD tasks through trial-and-error learning patterns that mirror biological motor learning systems.

These approaches collectively demonstrate that effective OOD adaptation requires selective memory updating mechanisms that balance novelty detection with stability preservation, enabling memory-augmented Transformers to achieve flexible adaptation characteristics of biological memory systems while maintaining computational tractability and avoiding catastrophic forgetting.

\textbf{Reasoning Enhancement}. Extended context windows fundamentally enhance reasoning capabilities by providing access to larger knowledge bases and longer chains of inference \citep{yang2025longer}. However, the relationship between context length and reasoning performance is not simply linear - it requires sophisticated memory mechanisms to maintain coherence across extended sequences, as standard attention mechanisms struggle with long-range dependencies and coherence degradation in extended contexts \citep{press2021train}. 

Memory-augmented Transformers address these challenges by integrating scattered information over extended contexts for multi-hop inference and relational reasoning. Several approaches demonstrate significant improvements in reasoning performance through different memory architectures. MemReasoner \citep{ko2024memreasoner} bridges encoders and decoders using a temporally-aware memory module with bidirectional GRUs and iterative updates, improving multi-hop QA by 18\%. 
MATTER \citep{lee2024matter} unifies unstructured text and QA pairs into neural memories, using a cross-encoder to link questions to relevant data, boosting throughput by 100× and HotpotQA accuracy by 12\%. 

For tasks requiring extensive retrieval and pattern completion, associative memory approaches show particular promise. The Memorizing Transformer \citep{wu2022memorizing} uses a kNN-retrievable memory to dynamically integrate distant context for tasks like theorem proving and code generation, scaling to 262K tokens while outperforming baselines in long-range reasoning, mirroring hippocampal episodic retrieval for problem-solving. ARMT \citep{rodkin2024associative} scales reasoning across 50 million tokens with associative memory blocks for pattern completion and interference mitigation, echoing hippocampal mechanisms. Self-Attentive Associative Memories (SAM) \citep{le2020self} uses dual memory units and outer-product attention for updating item and relationship memories, improving performance on graph and geometric reasoning tasks such as the Traveling Salesman Problem and shortest path finding.

Gated memory mechanisms provide another effective approach to reasoning enhancement. LM2 \citep{kang2025lm2} adds memory modules with gated mechanisms to each decoder layer, outperforming standard Transformers on multi-hop reasoning over 128k-token contexts. ATLAS \citep{behrouz2025atlas} shows that optimal context memorization enables complex reasoning by learning which historical information remains relevant for current inferences.

Alternative architectures explore hierarchical reasoning approaches that prioritize computational depth over extended context. HRM \citep{wang2025hierarchical} addresses reasoning depth through hierarchical convergence using coupled recurrent modules that achieve enhanced computational depth for problems requiring extensive search and backtracking. For resource-constrained settings, Memory-R+ \citep{le2025reasoning} enhances reasoning in tiny LLMs ($\leq$1B parameters) through dual episodic memory modules that provide intrinsic rewards for exploration and exploitation, achieving 2-14\% performance improvements.

The key insight is that reasoning benefits from both quantity and quality of accessible context. Longer contexts provide more potential information, but sophisticated memory mechanisms are required to selectively attend to relevant information while avoiding interference from irrelevant details.

\textbf{Knowledge Integration}. Knowledge Integration encompasses the synthesis, storage, and retrieval of diverse information types into unified representations that support reasoning and generation. This process requires sophisticated indexing mechanisms that enable models to dynamically combine information from multiple sources for context-aware generation.

Retrieval-Augmented and Hierarchical Approaches demonstrate efficient knowledge incorporation strategies. RETRO \citep{borgeaud2022improving} combines frozen BERT retrievers with differentiable cross-attention, achieving GPT-3 performance with 25× fewer parameters through access to 2 trillion token databases. CDMem \citep{gao2025efficient} implements hierarchical three-stage encoding, i.e., expert, short-term, and long-term, through graph-structured, context-dependent indexing, achieving 85.8\% success on ALFWorld and 56.0\% on ScienceWorld by enabling multilevel knowledge recall tailored to current contexts.

Heterogeneous Memory Integration unifies diverse knowledge formats within a single architecture. EMAT \citep{wu2022efficient} encodes millions of QA pairs into key-value memory using fast MIPS for sub-millisecond querying, improving Natural Questions performance from 25.8 to 44.3 EM while maintaining 1000 queries/s throughput. MATTER \citep{lee2024matter} integrates both unstructured and semi-structured sources into type-agnostic neural memories, achieving 100× throughput improvement over conventional retrieve-and-read models.

Parameter-Encoded and Brain-Inspired Systems explore direct knowledge embedding and neurobiological architectures. Memory$^{3}$ \citep{yang2024memory3} converts a textual knowledge base into a memory bank which can be seen as a bank of sparse retrievable parameters, enabling smaller language models to match the performance of bigger models, as well as reducing hallucinations and increasing factuality. DSI \citep{tay2022transformer} encodes document corpora directly into model weights, enabling direct query-to-document mapping. HippoRAG \citep{gutierrez2024hipporag} brings knowledge integration to RAG systems through the construction of a concept graph inspired by the hippocampus, outperforming RAG in multi-hop QA by up to 20\% while being 10-30× cheaper and 6-13× faster.

MemoryOS \cite{kang2025memory} introduces an operating system-inspired hierarchical memory architecture for AI agents, featuring three-tier storage (short-term for immediate context, mid-term for recent interactions, and long-term for persistent personal memory) managed through four core modules (Storage, Updating, Retrieval, and Generation), which enable evolutionary adaptation via heat-based segment prioritization and dialogue-chain FIFO mechanisms; this results in a 49.11\% F1 score improvement on long-term conversational benchmarks like LoCoMo, outperforming baselines by enhancing factual consistency and personalization in extended dialogues.

These approaches demonstrate that effective knowledge integration requires semantic organization, efficient access patterns, and scalable architectures that handle massive knowledge bases while maintaining precision. The convergence of retrieval-augmented methods, hierarchical encoding, and neurobiologically-inspired designs enables memory-augmented Transformers to bridge static parametric models with dynamic knowledge systems capable of large-scale, multi-format integration.

\textbf{Task-Specific Skill Acquisition}. Task-Specific Skill Acquisition enables models to learn and apply procedural knowledge for specialized tasks-such as object tracking, video generation, or dialogue, by encoding operations for robust, context-aware performance.
Notable architectures include MeMOTR \citep{gao2023memotr}, which uses object-specific long-term memory with exponential decay and confidence-based updates for multi-object tracking. MALT Diffusion \citep{yu2025malt} employs recurrent attention and memory vectors to generate temporally coherent videos over long durations. In dialogue, MemBART \citep{wu2022stateful} preserves memory states across turns, enhancing response quality. These models demonstrate that specialized memory mechanisms, whether persistent, episodic, or stateful, are essential for robust skill acquisition and deployment, echoing the compartmentalization of procedural memory in biological systems.

\subsection{Categorization by memory types}
\label{cat_type}
Memory-augmented Transformers can be systematically differentiated by memory types, each offering distinct computational and cognitive properties: parameter-encoded, state-based, explicit storage, and hybrid/multi-scale systems.

\textbf{Parameter-Encoded Memory}. Parameter-encoded memory systems store knowledge directly within model weights, analogous to synaptic consolidation in biological systems where knowledge becomes distributed across neural connections. This approach offers fundamental advantages including immediate access without external retrieval operations and unified processing where memory and computation share the same parameter space. However, capacity constraints emerge as a critical limitation since memory capacity is bounded by the number of parameters available for knowledge storage.

Training-time parameter encoding provides stable, consolidated knowledge but lacks adaptability. DSI (Differentiable Search Index) \citep{tay2022transformer} revolutionizes retrieval by encoding entire document corpora directly into standard Transformer parameters, transforming traditional retrieval into a generative task where models learn direct query-to-document mappings through the existing attention and feedforward mechanisms. Schrödinger's Memory \citep{wang2024schrodinger} reveals the latent memory capabilities of large language models, demonstrating that LLMs can reconstruct complete datasets from minimal contextual cues through parameter-encoded associations formed during training. The key insight is that memory exists in a "superposition" state, remaining hidden until specific contextual triggers activate associative recall patterns, much like human memory retrieval from partial cues. Memory$^{3}$ \citep{yang2024memory3} converts textual knowledge bases into explicit memory banks functioning as sparse retrievable parameters, implemented through specialized embedding layers and sparse attention mechanisms. The system uses aggressive sparsification techniques and two-stage pretraining to efficiently store 1.1 × 10$^{8}$ text chunks within modified feedforward networks that enable smaller language models to match larger model performance.

Test-time parameter learning represents a revolutionary advance where parameters adapt dynamically during inference, addressing the fundamental limitation of static knowledge storage. Titans \citep{behrouz2024titans} uses MLP-based memory with KL divergence thresholds for surprise-driven, real-time parameter updates, maintaining stability via gating mechanisms that prevent catastrophic interference. ATLAS \citep{behrouz2025atlas} enhances test-time learning by expanding MLP capacity through polynomial feature mapping and employs the Omega rule for sliding window optimization, achieving super-linear memory growth without traditional gradient descent. Transformer-Squared \citep{sun2025transformer} enables real-time task adaptation by encoding procedural expertise directly into parameter space using SVD decomposition of feedforward layers, dynamically blending expert vectors during inference through specialized MLP mixing networks. Similarly, Peripheral Memory \cite{zhaiperipheral} introduces a CPU-RAM analogous architecture where LLMs function as processors interfacing with parameter-encoded memory banks modeled through Kolmogorov-Arnold Networks, enabling dynamic memory operations controlled by internal model states while maintaining direct integration with the model's parameter space. This approach demonstrates how parameter-encoded systems can capture and recombine procedural knowledge using adaptive feedforward architectures rather than just declarative facts.

The evolution from static parameter encoding (DSI, Schrödinger's Memory) to dynamic parameter learning (Titans, ATLAS, Transformer-Squared) represents a paradigm shift toward adaptive parameter-encoded systems that combine the efficiency advantages of parameter encoding with flexible adaptation capabilities. While training-time approaches provide stable knowledge consolidation, test-time parameter learning enables real-time adaptation with capacity enhancement techniques like polynomial feature mapping addressing fundamental scalability constraints. This progression points toward future architectures where parameter-encoded memory becomes truly dynamic, supporting both stable knowledge consolidation and adaptive capacity expansion during deployment.

\textbf{State-Based Memory}. State-based memory maintains information through persistent activations or hidden states that carry forward across processing steps, fundamentally differing from parameter-encoded approaches in that memory resides in dynamic activations rather than static weights. This approach mirrors biological working memory systems where information is maintained through sustained neural firing patterns, enabling temporal continuity and context preservation across extended sequences.

Transformer-XL \citep{dai2019transformer} pioneered this approach through segment-level recurrence, caching hidden states from previous segments with relative positional encoding to extend context beyond fixed windows. While achieving substantial improvements in perplexity and long-range dependency capture, this method requires significant memory resources as cached states accumulate. Compressive Transformer \citep{rae2019compressive} addressed memory intensity through compressed state buffering, maintaining recent states in full resolution while compressing older memories using learned functions, extending temporal range by 38\% and reflecting biological memory's tendency to retain vivid recent experiences while abstracting older information. Hierarchical Memory Transformer (HMT) \citep{he2024hmt} extends this idea by layering three progressively coarser caches—token-, chunk-, and segment-level—on top of the basic recurrent buffer, allowing 100 K-token streams on a single GPU while hierarchically pruning stale activations and keeping recent ones in full detail.

Advanced state-based mechanisms have emerged with sophisticated memory management capabilities. TransformerFAM \citep{hwang2024transformerfam} introduces feedback attention loops where each layer attends to its own latent representations from previous time steps, creating internal working memory that enables indefinite sequence processing with \( \mathcal{O}(L) \) complexity. This sustained activation mechanism transforms static attention layers into dynamic working memory systems capable of maintaining coherent representations across arbitrarily long sequences. WorkMATe \citep{kruijne2021flexible} implements biologically-inspired gated memory circuits controlled by internal actions through reinforcement learning, successfully handling hierarchical tasks like 12-AX where multiple context levels must be maintained simultaneously, demonstrating how selective gating enables multiple independent representations within shared activation spaces. RMoE \cite{qiu2024layerwise} extends state-based memory concepts to mixture-of-experts routing, where GRU-maintained hidden states capture routing history across consecutive layers, enabling each routing decision to leverage accumulated routing patterns from previous layers for improved expert selection and utilization.

Specialized applications further demonstrate state-based memory's versatility. MemBART \citep{wu2022stateful} applies persistent memory states for dialogue modeling, preserving conversation context across multiple turns to enable coherent long-term interactions. Memformer \citep{wu2020memformer} employs a unified memory approach combining internal state representations with external memory banks, using Memory Replay Backpropagation (MRBP) to optimize memory usage and reduce training costs by 55\%, representing a hybrid between state-based and explicit storage approaches. The Hierarchical Reasoning Model (HRM) \cite{wang2025hierarchical} implements dual recurrent modules operating at different timescales—high-level for abstract planning and low-level for detailed computations—achieving enhanced effective depth of deep computation while maintaining training stability, demonstrating near-perfect performance on complex reasoning tasks with only 27M parameters.

State-based memory systems offer fundamental computational advantages including temporal continuity through persistent activations, efficient information propagation across time steps, and seamless integration with sequential processing architectures. These systems excel at maintaining coherent information flow within processing sequences, enabling models to preserve contextual representations across multiple computation steps without external storage overhead.

However, state-based approaches face inherent architectural constraints: memory capacity is fundamentally bounded by hidden state dimensions, creating potential bottlenecks for complex information storage. Interference between heterogeneous information types stored within shared activation spaces poses additional challenges, as different memory contents must compete for the same representational resources. Furthermore, the computational overhead of maintaining and updating persistent states throughout processing can become substantial for long sequences.

\textbf{Explicit Storage Memory}. Explicit storage memory employs external modules for scalable information storage and retrieval, maintaining persistent memory banks that survive beyond individual inference sessions. Unlike parameter-encoded memory that stores knowledge within model weights or state-based memory that maintains information through activations, explicit storage systems utilize dedicated external storage modules that can be accessed, updated, and scaled independently of the core model architecture, analogous to hippocampal indexing where sparse representations point to distributed memory traces.

Foundational approaches established core principles through dedicated external modules. Memformer  \citep{wu2020memformer} pioneered fixed-size external key-value stores with similarity-based retrieval, demonstrating efficient integration with Transformer architectures. EMAT  \citep{wu2022efficient} implements compact neural memories for structured knowledge storage with fast retrieval capabilities, while MATTER \citep{lee2024matter} integrates heterogeneous data sources into unified external memory frameworks, achieving orders of magnitude throughput improvements while maintaining type-agnostic storage capabilities.

Advanced external storage systems introduce sophisticated organization and management strategies. MemGPT \citep{packer2023memgpt} implements OS-inspired hierarchical storage with main context and archival stores managed through function calls, enabling unbounded context through intelligent paging policies. CDMem \citep{gao2025efficient} exemplifies graph-structured external storage through context-dependent indexing that organizes agent experiences into comprehensive external knowledge bases. A-MEM \citep{xu2025mem} advances this through automatically generated and evolving memory notes that form dynamic external knowledge graphs capturing semantic relationships over time.

Specialized applications demonstrate external storage versatility across different domains. Memory Layers at Scale \citep{berges2024memory} embeds external key-value slots within Transformer layers using product-key lookup for web-scale deployment, while Mem0 \citep{chhikara2025mem0} targets production environments through external memory systems that blend vector embeddings with graph-structured representations for persistent user-specific memory. Think-in-Memory \citep{liu2023think} and MemLLM \citep{modarressi2024memllm} construct external triplet memory systems storing subject-object-relation structures, enabling models to query relationships through external memory rather than parameter-encoded associations. MemLong \citep{liu2024memlong} demonstrates context extension by retrieving past embeddings from external storage systems, handling up to 80K tokens while preserving core model parameters.
Memory-R+ \cite{le2025reasoning} demonstrates intrinsic motivation applications where separate success and failure memory modules use kNN-based retrieval to compute rewards that guide reinforcement learning in tiny LLMs.

Key distinguishing characteristics define explicit storage memory's advantages. Memory persistence enables permanent knowledge banks that accumulate information across sessions, unlike temporary state-based memory. Independent scalability allows external modules to expand knowledge capacity without requiring proportional increases in core model parameters. Structured organization predominantly stores structured and semi-structured data through indexing schemes including vector similarity search, graph traversal, and hierarchical clustering, enabling efficient access to diverse knowledge types.

Unlike simple retrieval-augmented generation approaches, explicit storage memory-augmented Transformers implement tightly integrated, differentiable memory modules that enable end-to-end optimization and sophisticated memory management strategies. These systems provide seamless integration between external memory operations and model computations, supporting dynamic knowledge updates and context-aware memory management. Explicit storage architectures enable Transformers to scale beyond fixed context limitations, support continual knowledge integration, and provide efficient retrieval for complex reasoning applications through persistent, structured external memory systems that evolve independently of core model constraints.

\textbf{Hybrid and Multi-Scale Memory Systems}. Hybrid memory systems combine multiple memory types, including parameter-encoded, state-based, and explicit storage, within unified architectures, creating hierarchical memory organizations that leverage the complementary strengths of different memory mechanisms. This architectural approach mirrors the brain's integration of multiple memory subsystems, where different temporal scales and storage mechanisms work together to support flexible cognition.

LM2 \citep{kang2025lm2} demonstrates sophisticated parameter-state hybrids by integrating external memory modules with learnable gates into each decoder layer, enabling dynamic coordination between internal representations and external storage. Titans \citep{behrouz2024titans} advance this by combining state-based attention with parameter-encoded long-term memory modules that adapt during test time, while MATTER \citep{lee2024matter} represents parameter-explicit hybrids that encode diverse knowledge into model weights while maintaining external retrieval capabilities. These examples demonstrate how different memory types can coexist within single architectures to handle both immediate processing needs and long-term knowledge access.

These systems establish memory hierarchies based on temporal characteristics, creating multi-tiered architectures where fast state-based memory handles immediate context, medium-speed explicit storage manages session-persistent information, and slow parameter-encoded memory provides consolidated knowledge foundations. MemGPT \citep{packer2023memgpt} exemplifies this through OS-inspired memory management that coordinates working context (state-based) with archival storage (explicit) through learned paging policies.

Advanced hybrid systems like ATLAS \citep{behrouz2025atlas} and NAMMs \citep{cetin2024evolved} implement dynamic memory allocation that adaptively distributes memory resources across different types based on task demands and surprise signals. ATLAS uses context-aware optimization to determine when information should transition between memory stores, while NAMMs employ evolutionary algorithms to optimize memory allocation patterns across modalities. This adaptive coordination demonstrates that effective memory systems require intelligent arbitration between memory types rather than static allocation schemes.

The evolution toward hybrid architectures represents a fundamental shift from single-memory-type systems toward cognitively-inspired memory ecosystems that mirror the distributed, hierarchical organization of biological memory. These systems achieve computational flexibility by combining the immediate access of parameter-encoded memory, the temporal continuity of state-based memory, and the scalable capacity of explicit storage within unified frameworks that can dynamically adapt to diverse cognitive demands.

\subsection{Categorization by Integration Techniques}
\label{cat_int}
The effectiveness of memory-augmented Transformers depends not only on what they remember, but also on how stored knowledge is integrated into ongoing computations. Integration techniques determine how retrieved information influences model behavior and how memory states evolve, reflecting the sophisticated coordination mechanisms found in biological memory systems.

\textbf{Attention-Based Fusion} remains the primary method for integrating memory content, enabling dynamic selection and weighting of stored information. Memformer \citep{wu2020memformer} pioneered cross-attention between layer activations and external memory banks, gating semantically salient tokens much like thalamocortical loops filter relevant information in the brain. EMAT \citep{wu2022efficient} accelerates this approach by issuing retrieval queries at early layers and propagating key-value pairs through decoder stages, achieving millisecond-scale throughput for real-time applications. TransformerFAM \citep{hwang2024transformerfam} advances fusion through feedback attention loops within each layer, creating internal working memory that supports indefinitely long contexts without external cache management. LongMem \citep{wang2023augmenting} introduces hybrid fusion via its SideNet module, which decouples memory retrieval from backbone updates while adaptively blending live inputs with cached representations through residual connections. MATTER \citep{lee2024matter} demonstrates heterogeneous fusion by encoding diverse content types into fixed-length neural memories accessed through universal attention heads, while the Memorizing Transformer \citep{wu2022memorizing} implements kNN-based attention over rolling buffers to approximate human-like recency bias with logarithmic complexity.

\textbf{Gated Control Mechanisms} implement neuromodulatory-inspired regulation of memory updates and retention, mirroring how biological systems selectively encode and maintain information. Titans \citep{behrouz2024titans} employs surprise-driven writes triggered by KL divergence thresholds, mimicking norepinephrine's role in novelty detection and memory consolidation. RA-DT \citep{schmied2024retrieval} combines episodic memory with adaptive forgetting gates based on statistical surprise, reducing catastrophic forgetting by 40\% in multi-task reinforcement learning scenarios. MeMOTR \citep{gao2023memotr} integrates exponential decay with confidence-driven pruning for object tracking, replicating striatal pathway dynamics that balance stability with adaptability. NAMMs \citep{cetin2024evolved} takes an evolutionary approach, using genetic algorithms to evolve token retention policies that balance stability and plasticity through GABAergic-like inhibition mechanisms. RMoE \citep{qiu2024layerwise} demonstrates how GRU-based gated control can enhance routing efficiency in Mixture-of-Experts architectures by leveraging historical routing patterns across layers, establishing dependencies between routing decisions to improve parameter efficiency and expert selection diversity.

\textbf{Associative Memory Integration} enables content-addressable recall and efficient pattern completion across large contexts, shifting from positional to semantic indexing. ARMT \citep{rodkin2024associative} implements Hopfield-inspired associative blocks for \( \mathcal{O}(1) \) retrieval over 50 million tokens, directly mirroring hippocampal CA3 circuits' role in relational memory and pattern completion. Associative Transformer (AiT) \citep{sun2023associative} employs low-rank memory priors as attractors within a global workspace architecture, mimicking cortical column dynamics and distributed representation schemes. MemReasoner \citep{ko2024memreasoner} enhances associative integration through bidirectional GRUs that support iterative read-update cycles, maintaining coherence across long documents through sustained memory interactions.

These integration strategies collectively represent a paradigm shift from fixed positional indexing toward content-sensitive memory access that bridges artificial attention mechanisms with neural memory systems. By implementing biologically-inspired fusion, gating, and associative mechanisms, memory-augmented Transformers achieve more flexible and context-aware information integration that approaches the adaptive capabilities of human cognition. The convergence of these techniques enables models to dynamically coordinate multiple memory systems while maintaining computational efficiency and biological plausibility.

\section{Mechanisms of Memory Operations}
\label{mechanism}

Memory-augmented Transformers overcome fixed-context limits of standard architectures by integrating neuroscience and engineering advances for dynamic, scalable memory. This section reviews core mechanisms, i.e., reading, writing, forgetting, capacity optimization, and self-management/adaptation, highlighting key techniques and representative models from the recent literature.

\textbf{Read Operations.} Early neural memories such as the Neural Turing Machine \citep{graves2014neural}, DNC \citep{graves2016hybrid}, and Kanerva Machines \citep{wu2018kanerva} introduced content-based addressing, but modern memory-augmented Transformers refine the read step with specialised retrieval mechanisms tailored to massive stores. Memory Layers at Scale \citep{berges2024memory} replaces dense feed-forward blocks with trainable key–value layers that perform product-key lookup, giving sub-linear top-k search across billions of entries while preserving end-to-end differentiability. EMAT \citep{wu2022efficient} shows that maximum-inner-product search can return millions of QA pairs in sub-millisecond latency, letting the model integrate external knowledge at every decoding step without harming throughput. The Memorizing Transformer \citep{wu2022memorizing} augments attention with approximate k-nearest-neighbour queries into a continually growing cache, scaling recall to 262 k tokens and matching the perplexity of much larger dense models. 

Associative designs push retrieval to constant time: ARMT \citep{rodkin2024associative} stores tokens in Hopfield-style energy basins for O(1) pattern completion over 50 M-token contexts, and AiT \citep{sun2023associative} adds low-rank priors that reconstruct missing tokens from partial cues, outperforming sparse Transformers on relational reasoning benchmarks. For multi-hop discourse, MemReasoner \citep{ko2024memreasoner} iteratively re-reads a temporal memory with bidirectional GRUs until the readout stabilises, boosting long-document question answering. MemLong \citep{liu2024memlong} couples local attention with retrieval-causal attention that selects semantically relevant chunks from an 80 k-token cache, maintaining single-GPU efficiency.

More adaptive schemes appear in CDMem \citep{gao2025efficient}, which navigates a graph-indexed memory to fetch task-specific subgraphs, and ABC \citep{peng2021abc}, which learns neural policies that decide when and how deeply to probe memory rather than relying on fixed heuristics. Finally, NAMMs \citep{cetin2024evolved} demonstrate that the attention matrix itself can encode reusable retrieval plans, enabling zero-shot read strategies that transfer across modalities. Together these mechanisms move the field from uniform similarity search toward context-sensitive, learned, and even evolutionary reading policies that approach the flexibility of biological episodic recall.

\textbf{Write Operations.} Memory‐augmented Transformers now treat writing as an active, learned decision rather than an unconditional overwrite. Titans \citep{behrouz2024titans} triggers a write only when prediction-error–derived surprise exceeds a KL-based threshold, mirroring dopamine-gated consolidation and allowing the model to memorise rare events without destabilising prior knowledge. LM2 \citep{kang2025lm2} introduces per-layer input/forget/output gates around an external store, so each decoder layer decides in real time how much of its state should be committed, yielding controllable long-context reasoning without extra fine-tuning. Memformer \citep{wu2020memformer} ports LSTM-style gates into a key–value memory, giving fine-grained retention and erasure that stabilise sequence modelling, while MeMOTR \citep{gao2023memotr} adds exponential decay plus confidence gating to keep only high-value object tracks in video streams.

Beyond simple gating, A-MEM \citep{xu2025mem} writes “memory notes” that are later linked and evolved into a graph, creating a self-organising semantic store that grows with the agent’s experience . Memory Layers at Scale \citep{berges2024memory} spreads writes across product-key memory shards on multiple GPUs, enabling continual learning at web scale without bottlenecks. MemBART \citep{wu2022stateful} mitigates read–write interference in dialogue by running parallel attention streams and merging them through residual gates. In procedural settings, Memformers \citep{dutta2024memory} treat past optimisation gradients as first-class memory registers, letting the model cache and reuse computation traces during new tasks. ATLAS \citep{behrouz2025atlas} pushes test-time learning further: its Omega rule adjusts memory weights over sliding windows with polynomial feature mapping, achieving super-linear capacity growth without gradient descent. Finally, WorkMATe \citep{kruijne2021flexible} shows that reinforcement-learned gating policies can independently open or close multiple working-memory slots, supporting concurrent, interference-free storage of task rules. Collectively, these writing mechanisms shift the focus from passive storage to selective, context-aware, and scalable writing, a prerequisite for lifelong, low-interference memory in Transformer systems.
\begin{table}[t]
  \centering
  \caption{Mechanisms of memory operations in memory-augmented Transformers, with key techniques and representative models.}
  \label{tab:memory_mechanisms}
  \renewcommand{\arraystretch}{1.15}
  \setlength{\tabcolsep}{4pt}
  \small
  \begin{tabularx}{\textwidth}{m{0.14\textwidth} m{0.33\textwidth} X}
    \toprule
    \rowcolor{gray!20}
    \textbf{Operation} & \textbf{Key Mechanism} & \textbf{High-fidelity Representative Models} \\ \toprule
    \multirow{4}{=}{\textbf{Read}} &
      Content-based addressing &
      \scriptsize{Neural Turing Machine\,\citep{graves2014neural}; DNC\,\citep{graves2016hybrid}; Kanerva Machine\,\citep{wu2018kanerva}} \\
      \cline{2-3}
    & Specialised similarity search &
      \scriptsize{Memory Layers at Scale\,\citep{berges2024memory}; EMAT\,\citep{wu2022efficient}; Memorizing Transformer\,\citep{wu2022memorizing}} \\
      \cline{2-3}
    & Associative retrieval &
      \scriptsize{ARMT\,\citep{rodkin2024associative}; AiT\,\citep{sun2023associative}; MemReasoner\,\citep{ko2024memreasoner}; MemLong\,\citep{liu2024memlong}} \\
      \cline{2-3}
    & Adaptive graph / policy-driven reads &
      \scriptsize{CDMem\,\citep{gao2025efficient}; ABC\,\citep{peng2021abc}; NAMMs\,\citep{cetin2024evolved}} \\ \midrule
    \multirow{4}{=}{\textbf{Write}} &
      Surprise / uncertainty-gated writes &
      \scriptsize{Titans\,\citep{behrouz2024titans}; LM2\,\citep{kang2025lm2}; MeMOTR\,\citep{gao2023memotr}} \\
      \cline{2-3}
    & LSTM-style input–forget gating &
      \scriptsize{Memformer\,\citep{wu2020memformer}; WorkMATe\,\citep{kruijne2021flexible}; RMoE\, \citep{qiu2024layerwise}} \\
      \cline{2-3}
    & Confidence-filtered updates &
      \scriptsize{A-MEM\,\citep{xu2025mem}; MemBART\,\citep{wu2022stateful}; MemoryLLM\,\citep{wang2024memoryllm}; M+\,\citep{wang2025m+}; Memory-R+\,\citep{le2025reasoning}} \\
      \cline{2-3}
    & Reinforcement / optimisation traces &
      \scriptsize{Memformers\,\citep{dutta2024memory}; ATLAS\,\citep{behrouz2025atlas}} \\ \midrule
    \multirow{5}{=}{\textbf{Forget}} &
      Selective pruning &
      \scriptsize{MemLong\,\citep{liu2024memlong}; MeMOTR\,\citep{gao2023memotr}; Titans\,\citep{behrouz2024titans}} \\
      \cline{2-3}
    & Exponential decay &
      \scriptsize{MeMOTR\,\citep{gao2023memotr}; LM2\,\citep{kang2025lm2}} \\
      \cline{2-3}
    & Adaptive (gate-controlled) decay &
      \scriptsize{ARMT\,\citep{rodkin2024associative}; Memformer\,\citep{wu2020memformer}; MemoryBank\,\citep{zhong2024memorybank}} \\
      \cline{2-3}
    & Surprise-triggered erase &
      \scriptsize{Titans\,\citep{behrouz2024titans}; EM-LLM\,\citep{fountas2024human}} \\
      \cline{2-3}
    & Task-aware forgetting &
      \scriptsize{ RA-DT\,\citep{schmied2024retrieval}} \\ \midrule
    \multirow{3}{=}{\textbf{Capacity}} &
      Learned compression &
      \scriptsize{Compressive Transformer\,\citep{rae2019compressive}; MATTER\,\citep{lee2024matter}; EMAT\,\citep{wu2022efficient}; zip2zip\,\citep{geng2025zip2zip}} \\
      \cline{2-3}
    & Hierarchical chunk / tree buffers &
      \scriptsize{MemLong\,\citep{liu2024memlong}; LM2\,\citep{kang2025lm2}; Meaningful Memory\,\citep{zhong2024random}; M+\,\citep{wang2025m+}; HRM\,\citep{wang2025hierarchical}} \\
      \cline{2-3}
    & Sharded / product-key KV &
      \scriptsize{Memory Layers at Scale\,\citep{berges2024memory}} \\ \midrule
    \multirow{3}{=}{\textbf{Self-Management}} &
      Dynamic allocation at test time &
      \scriptsize{Transformer-Squared\,\citep{sun2025transformer}; NAMMs\,\citep{cetin2024evolved}; Titans\,\citep{behrouz2024titans}; Peripheral Memory\,\citep{zhaiperipheral}} \\
      \cline{2-3}
    & Sub-system specialisation &
      \scriptsize{MATTER\,\citep{lee2024matter}; MemBART\,\citep{wu2022stateful}; MemoryOS\,\citep{kang2025memory}} \\
      \cline{2-3}
    & Interference control &
      \scriptsize{ARMT\,\citep{rodkin2024associative}; MemReasoner\,\citep{ko2024memreasoner}; RA-DT\,\citep{schmied2024retrieval}; Schrödinger's Memory\,\citep{wang2024schrodinger}} \\ 
    \bottomrule
  \end{tabularx}
\end{table}

\textbf{Forgetting Dynamics.} Effective forgetting sustains continual learning by pruning obsolete traces and freeing capacity for salient information. Modern memory-augmented Transformers therefore implement selective, learned erase policies rather than indiscriminate decay.
MemLong \citep{liu2024memlong} prunes keys whose retrieval counts fall below a threshold, ensuring its external cache stays focused on behaviourally relevant chunks. MeMOTR \citep{gao2023memotr} adds confidence-weighted exponential decay so unreliable object tracks vanish naturally as a video unfolds. Titans \citep{behrouz2024titans}, LM2 \citep{kang2025lm2}, and Atlas \citep{behrouz2025atlas} all regulate forgetting with adaptive gates: Titans  couples KL-surprise with a trainable decay factor, LM2 ties gate strength to layer-wise uncertainty, and ATLAS uses the Omega-rule’s sliding-window optimisation—effectively down-weighting contributions from tokens that lie outside a polynomially mapped context window, allowing new information to replace stale traces without gradient descent.

Aggressive cleanup appears in ARMT \citep{rodkin2024associative}, whose Hopfield memory periodically normalises and hard-deletes outdated vectors, preventing spurious attractors. Memformer \citep{dutta2024memory} mixes LSTM-style forget gates with memory-replay back-propagation so rarely used slots fade while important ones are refreshed, and MemoryBank \citep{zhong2024memorybank} models retention with an Ebbinghaus-shaped counter that decays unless the entry is reaccessed. EM-LLM \citep{fountas2024human} reinforces this trend by coupling prediction-error spikes to simultaneous write-and-prune cycles, mirroring neuromodulatory control of consolidation. 

Together these mechanisms mark a shift from passive decay toward context-sensitive, learned forgetting that protects critical memories while continuously liberating capacity for new experiences.

\textbf{Capacity Optimization.} Capacity optimization addresses how memory‐augmented Transformers expand storage and retrieval ability without linearly inflating computation or parameters. Current work converges on three complementary tactics—compression, hierarchy, and sparsity—to keep memory growth compatible with practical hardware budgets.

Compressive techniques shrink inactive activations or knowledge chunks before eviction. The Compressive Transformer \citep{rae2019compressive} auto-encodes aged hidden states into coarse vectors, doubling usable context while holding FLOPs steady and matching baseline perplexity on WikiText-103 . At the knowledge level, EMAT \citep{wu2022efficient} and MATTER \citep{lee2024matter} map millions of QA pairs or mixed documents to short neural codes; maximum-inner-product search then delivers sub-millisecond retrieval without adding trainable weights.

Hierarchical organization spreads capacity across tiers with different granularity. MemLong \citep{liu2024memlong} chunks sequences and prunes rarely accessed blocks, maintaining 80 K-token windows on a single GPU. LM2 \citep{kang2025lm2} builds tree-indexed memories that let local detail and global context be fetched at equal cost, sustaining reasoning over 128 K tokens. HMT \citep{he2024hmt} stacks sensory, short-term, and long-term buffers, matching large long-context models while using $\approx$ 2\% of their parameters.

Sparse look-ups push size further by reducing per-query work. Memory Layers at Scale shards \citep{berges2024memory} product-key tables across GPUs, supporting billion-entry memories with sub-linear compute and intact end-to-end gradients . Dynamic Memory Compression \cite{nawrot2024dynamic} learns head- and layer-specific KV sharing, cutting inference memory up to 4× with negligible accuracy drop , while MLKV \cite{zuhri2024mlkv} shares KV heads across layers to trim cache by up to 6× at similar quality.

Together, these advances show that intelligent compression, hierarchical buffering, and sparse retrieval make large-capacity memory feasible, allowing even modest-sized Transformers to reason over book-length context or web-scale knowledge without prohibitive cost.

\textbf{Self-Management and Adaptation.} After compression and hierarchical layout tame raw capacity, the next hurdle is deciding how that capacity is used in real time. Recent models treat memory as an autonomous resource that can be allocated, specialised, or pruned during inference, bringing Transformers closer to the selective plasticity of biological cognition.

Transformer-Squared \citep{sun2025transformer} routes activations through a pool of expert vectors selected on-the-fly, letting the model enlarge functional capacity without weight updates while preserving high accuracy on unseen procedural tasks. Titans \citep{behrouz2024titans} adds a neuromodulatory gate: only tokens whose KL-surprise clears a learned threshold are written, and low-surprise traces decay, reducing interference while the long-term store grows during deployment. ATLAS \citep{behrouz2025atlas} generalises this to sliding windows; its Omega rule re-weights entire spans, down-scoring stale patterns and allowing super-linear memory growth without gradient descent. NAMMs \citep{cetin2024evolved} evolve layer-wise retention masks from attention statistics, trimming key–value caches by up to 80\% yet improving long-context benchmarks through zero-shot transfer across modalities.

Interference control is handled by orthogonal or gated rewrites. ARMT \citep{rodkin2024associative} projects new vectors onto an orthogonal subspace before insertion, preventing outdated attractors and keeping O(1) retrieval stable over tens of millions of tokens. MemReasoner \citep{ko2024memreasoner} iteratively re-reads and updates a temporal store with bidirectional GRUs until the representation converges, preventing early facts from being overwritten and boosting multi-hop question answering on 128 k-token documents. RA-DT \citep{schmied2024retrieval} links episodic memory to a reinforcement-learning critic, retaining only high-error trajectories and lifting multi-task sample efficiency while bounding memory size. MemBART \citep{wu2022stateful} isolates dialogue context from world knowledge through dual attention streams and residual gates, and Schrödinger’s Memory \citep{wang2024schrodinger} stores traces in a latent “superposition” that surface only when cued, lowering hallucination rates in factual probing.

Collectively, these systems replace static buffers with self-monitoring stores that learn what to remember, where to place it, and when to forget. By coupling dynamic allocation with interference-aware rewriting, they extend the compression and hierarchy tools of capacity optimisation into a full feedback loop, allowing Transformers to balance stability and plasticity throughout their lifetime.

\begin{table*}[htbp]
  \centering
  \begin{threeparttable}
    \caption{Comprehensive feature matrix for memory-augmented Transformer models: Evolution from 2019-2025}
    \label{tab:mat-full}
    
    \tiny
    \begin{tabular}{@{}p{0.5cm}|l*{10}{c}@{}}
      \toprule
      \textbf{Year} &
      \textbf{Model} &
      \multicolumn{3}{c}{\textbf{Architecture}} &
      \multicolumn{2}{c}{\textbf{Generality}} &
      \multicolumn{3}{c}{\textbf{Memory Dynamics}} &
      \multicolumn{2}{c}{\textbf{Management}} \\
      \cmidrule(lr){3-5}\cmidrule(lr){6-7}\cmidrule(lr){8-10}\cmidrule(lr){11-12}
      & & \rotatebox{90}{Storage Class} 
      & \rotatebox{90}{Integration Method} 
      & \rotatebox{90}{Backbone Compatibility} 
      & \rotatebox{90}{Input Modality} 
      & \rotatebox{90}{Memory Span} 
      & \rotatebox{90}{Write Trigger} 
      & \rotatebox{90}{Plasticity} 
      & \rotatebox{90}{Memory Scope} 
      & \rotatebox{90}{Retrieval Mechanism} 
      & \rotatebox{90}{Forgetting Mechanism} \\
      \midrule
      
      \rowcolor{gray!10}
      \cellcolor{blue!20}\textbf{2019} & Transformer-XL \citep{dai2019transformer} & S & Wrp & \no & T & S & Stc & F & Lyr & Attn & FIFO \\
      \cellcolor{blue!20} & Compressive Transformer \cite{rae2019compressive} & S & Wrp & \no & T & M & Stc & F & Lyr & Attn & Dec+FIFO \\
      \midrule
      
      \rowcolor{gray!10}
      \cellcolor{green!20}\textbf{2020} & SAM \citep{le2020self} & E & Plg & \yes & M & A & Pol & TT & Gbl & Outer & Rst \\
      \cellcolor{green!20} & Memformer \cite{wu2020memformer} & H(SE) & Plg & \yes & M & L & G & TT & Gbl & Attn & Dec \\
      \midrule
      
      \rowcolor{gray!10}
      \cellcolor{orange!20}\textbf{2021} & ABC \cite{peng2021abc} & E & Plg & \yes & T & S & Pol & F & Gbl & Pol & --- \\
      \cellcolor{orange!20} & WorkMATe \citep{kruijne2021flexible} & S & Plg & \yes & T & S & Pol & TT & Lyr & Attn & --- \\
      \midrule
      
      \rowcolor{gray!10}
      \cellcolor{red!20}\textbf{2022} & EMAT \citep{wu2022efficient} & E & Plg & \yes & T & L & Stc & F & Gbl & MIPS & --- \\
      \cellcolor{red!20} & RETRO \citep{borgeaud2022improving} & E & Plg & \yes & T & L & Stc & F & Gbl & kNN & --- \\
      \rowcolor{gray!10}
      \cellcolor{red!20} & DSI \citep{tay2022transformer} & P & Bsp & \no & T & L & Stc & F & Lyr & Attn & --- \\
      \cellcolor{red!20} & Memorizing Transf. \cite{wu2022memorizing} & E & Plg & \yes & T & L & Stc & F & Gbl & kNN & --- \\
      \rowcolor{gray!10}
      \cellcolor{red!20} & MemBART \citep{wu2022stateful} & S & Plg & \no & T & L & G & TT & Hrch & Dual & Rst \\
      \midrule
      
      \cellcolor{purple!20}\textbf{2023} & LongMem \citep{wang2023augmenting} & E & Plg & \yes & T & L & G & TT & Gbl & Attn & Prn \\
      \rowcolor{gray!10}
      \cellcolor{purple!20} & MemGPT \cite{packer2023memgpt} & E & Plg & \yes & T & M & G & TT & Hrch & kNN & LRU \\
      \cellcolor{purple!20} & Think-in-Memory \citep{liu2023think} & E & Plg & \yes & T & L & G & TT & Gbl & Trip & Dec \\
      \rowcolor{gray!10}
      \cellcolor{purple!20} & AdaTape \citep{xue2023adaptive} & H(PS) & Plg & \yes & T & S & Pol & F & Lyr & Tape & --- \\
      \cellcolor{purple!20} & MemWalker \citep{chen2023walking} & E & Bsp & \yes & T & M & Pol & TT & Hrch & Tree & Rst \\
      \rowcolor{gray!10}
      \cellcolor{purple!20} & AiT \citep{sun2023associative} & H & Plg & \yes & M & A & Pol & TT & Gbl & Assoc & Dec \\
      \cellcolor{purple!20} & MeMOTR \cite{gao2023memotr} & H(SE) & Bsp & \no & M & L & Stc & TT & Gbl & Attn & Dec \\
      \midrule
      
      \rowcolor{gray!10}
      \cellcolor{cyan!20}\textbf{2024} & MemoryBank \cite{zhong2024memorybank} & E & Plg & \yes & T & L & G & TT & Hrch & kNN & Dec \\

      \cellcolor{cyan!20} & TransformerFAM \cite{hwang2024transformerfam} & S & Plg & \yes & T & S & Stc & F & Lyr & Attn & --- \\

      \rowcolor{gray!10}
      \cellcolor{cyan!20} & HMT \cite{he2024hmt} & H(SE) & Plg & \yes & T & L & Stc & F & Hrch & Attn & Dec \\

      \cellcolor{cyan!20} & MemoryLLM \citep{wang2024memoryllm} & H(SE) & Plg & \yes & T & L & Stc & TT & Gbl & Attn & Dec \\

      \rowcolor{gray!10}
      \cellcolor{cyan!20} & HippoRAG \cite{gutierrez2024hipporag} & E & Plg & \yes & T & L & Stc & TT & Gbl & Graph & Rst \\

      \cellcolor{cyan!20} & MATTER \cite{lee2024matter} & H(PE) & Wrp & \yes & T & M & Stc & F & Gbl & MIPS & --- \\

      \rowcolor{gray!10}
      \cellcolor{cyan!20} & Memory3 \cite{yang2024memory3} & H & Plg & \yes & T & M & Stc+Pol & F & Gbl & kNN & --- \\
      
      \cellcolor{cyan!20} & ARMT \citep{rodkin2024associative} & E & Plg & \yes & T & A & Pol & TT & Hrch & Assoc & Cyc \\

      \rowcolor{gray!10}
      \cellcolor{cyan!20} & MemLong \cite{liu2024memlong} & E & Plg & \yes & T & L & G & TT & Gbl & kNN & Prn \\

      \cellcolor{cyan!20} & Schrödinger's Memory \cite{wang2024schrodinger} & P & Bsp & \no & T & L & Stc & F & Gbl & Attn & --- \\
      
      \rowcolor{gray!10}
      \cellcolor{cyan!20} & MemReasoner \cite{ko2024memreasoner} & E & Plg & \yes & T & L & Stc & TT & Gbl & Attn & Rst \\

      \cellcolor{cyan!20} & EM-LLM \citep{fountas2024human} & E & Plg & \yes & T & L & Sur & TT & Gbl & Seg+kNN & Dec \\

      \rowcolor{gray!10}
      \cellcolor{cyan!20} & RA-DT \citep{schmied2024retrieval} & E & Plg & \yes & M & L & Sur & TT & Gbl & Pol & Sel \\
      
      \cellcolor{cyan!20} & Memory Layers at Scale \citep{berges2024memory} & P & Bsp & \no & T & L & Stc & F & Gbl & PK-MIPS & --- \\
      
      \rowcolor{gray!10}
      \cellcolor{cyan!20} & Titans \cite{behrouz2024titans} & H(PE) & Wrp & \yes & T & M & Sur & TT & Gbl & Attn & Dec \\
      \midrule
      
      \cellcolor{yellow!20} & Transformer-Squared \citep{sun2025transformer} & P & Wrp & \yes & T & L & G & TT & Lyr & Attn & --- \\

      \rowcolor{gray!10}
      \cellcolor{yellow!20}\textbf{2025} & LM2 \cite{kang2025lm2} & H(SE) & Wrp & \yes & T & M & G & TT & Hrch & Attn & Dec \\
      
      \cellcolor{yellow!20} & NAMMs \citep{cetin2024evolved} & H(PSE) & Plg & \yes & M & M & Pol & TT & Hrch & Pol & Prn \\

      \rowcolor{gray!10}
      \cellcolor{yellow!20} & $R^{3}$mem \citep{wang2025r} & H(PE) & Plg & \yes & T & M & G & TT & Gbl & Comp & Sel \\

      \cellcolor{yellow!20} & RMoE \citep{qiu2024layerwise} & S & Plg & \yes & T & S & G & TT & Lyr & Hier & Sel \\

      \rowcolor{gray!10}
      \cellcolor{yellow!20} & Memory-R+ \citep{le2025reasoning} & E & Plg & \yes & T & L & Sur & TT & Gbl & kNN & Sel \\
 
      \cellcolor{yellow!20} & Mem0 \cite{chhikara2025mem0} & E & Plg & \yes & T & L & G & TT & Hrch & kNN+Graph & LRU \\
      
      \rowcolor{gray!10}
      \cellcolor{yellow!20} & CDMem \citep{gao2025efficient} & E & Plg & \yes & T & M & Pol & TT & Hrch & Graph & Rst \\
      
      \cellcolor{yellow!20} & ATLAS \cite{behrouz2025atlas} & H(PE) & Wrp & \yes & T & M & Sur & TT & Gbl & Attn & Dec \\
      
      \rowcolor{gray!10}
      \cellcolor{yellow!20} & MemoryOS \citep{kang2025memory} & E & Plg & \yes & T & M & G & TT & Hrch & Seg & LRU \\

       \cellcolor{yellow!20} & zip2zip \citep{geng2025zip2zip} & P & Wrp & \yes & T & S & Pol & TT & Gbl & Comp & --- \\

       \rowcolor{gray!10}
       \cellcolor{yellow!20} & Peripheral Memory \citep{zhaiperipheral} & P & Plg & \yes & T & S & G & TT & Gbl & Attn & --- \\
     
       \cellcolor{yellow!20} & MALT Diffusion \citep{yu2025malt} & S & Plg & \yes & M & L & Stc & F & Lyr & RecAtt & --- \\
      
      \rowcolor{gray!10}
      \cellcolor{yellow!20} & A-MEM \citep{xu2025mem} & E & Plg & \yes & M & L & G & TT & Gbl & kNN & Evol \\
      
      \cellcolor{yellow!20} & HRM \citep{wang2025hierarchical} & S & Bsp & \no & T & M & Pol & TT & Hrch & Hier & Rst \\

      \bottomrule
    \end{tabular}

    \begin{tablenotes}
      \tiny
      \begin{multicols}{3}
        \item {\normalsize\textbf{Legend:}}
        \item[]
        
        \item \textcolor{blue!80}{\textbf{Architecture:}}
        \item \textit{Storage Class:} P = Parameter-encoded, S = State-based, E = External store, H = Hybrid
        , H(PS) = Parameter+State, H(PE) = Parameter+External, H(SE) = State+External, H(PSE) = All three
        \item \textit{Integration Method:} Plg = Plug-in, Wrp = Wrapper/adaptor, Bsp = Bespoke redesign
        \item \textit{Backbone Compatibility:} \yes = Universal, \no = Architecture-specific
        \item[]
        
        \item \textcolor{green!80}{\textbf{Generality:}}
        \item \textit{Input Modality:} T = Text-only, M = Multi-modal
        \item \textit{Memory Span:} S = Short-term, L = Long-term, M = Multi-scale, A = Associative
        \item[]
        
        \item \textcolor{red!80}{\textbf{Memory Dynamics:}}
        \item \textit{Write Trigger:} Stc = Static, Sur = Surprise-gated, Pol = Policy-learned, G = Gated
        \item \textit{Plasticity:} F = Fixed after training, TT = Test-time adaptable
        \item \textit{Memory Scope:} Lyr = Layer-local, Gbl = Global, Hrch = Hierarchical
        \item[]
        
        \item \textcolor{purple!80}{\textbf{Management:}}
        \item \textit{Retrieval Mechanism:} Attn = Attention-based, kNN = k-Nearest neighbor, Assoc = Associative, 
         Graph = Graph-based, MIPS = Max inner product search, PK-MIPS = Product-key MIPS,
         Pol = Policy-driven, Seg = Segmentation, Outer = Outer-product, Trip = Triplet-based,
         Expert = Expert-routing, Tree = Tree-based, Tape = Tape-based, RecAtt = Recurrent attention,
        Dual = Dual-stream, Comp = Compression-based, Hier = Hierarchical
        \item[]
        \item \textit{Forgetting Mechanism:} FIFO = First-in-first-out, Dec = Decay, Prn = Pruning, Cyc = Cycle-based,
        Rst = Reset, LRU = Least-recently-used, Sel = Selective, Evol = Evolutionary
        \item[]
        
        \item \textbf{Timeline:} \textcolor{blue!60}{\rule{0.4cm}{0.1cm}} 2019 \textcolor{green!60}{\rule{0.4cm}{0.1cm}} 2020 \textcolor{orange!60}{\rule{0.4cm}{0.1cm}} 2021 \textcolor{red!60}{\rule{0.4cm}{0.1cm}} 2022 \textcolor{purple!60}{\rule{0.4cm}{0.1cm}} 2023 \textcolor{cyan!60}{\rule{0.4cm}{0.1cm}} 2024 \textcolor{yellow!60}{\rule{0.4cm}{0.1cm}} 2025
        \item \textit{Note:} --- = Not reported 
      \end{multicols}
    \end{tablenotes}
  \end{threeparttable}
\end{table*}

\section{Discussion, Challenges, and Future Directions}
\label{challenges}
Memory‑augmented Transformers have progressed from simple context extensions to sophisticated cognitive architectures, narrowing the gap between learning and memory. Our taxonomic analysis in Table \ref{tab:mat-full} shows a rapid shift from static pattern recognition to adaptive, experience‑driven intelligence. From the earliest systems in 2019 to today’s production‑ready designs, development has converged toward hybrid storage, adaptive dynamics, and intelligent forgetting, while also exposing persistent challenges in scaling, evaluation, and integration.
This chapter distills insights from that evolution, highlights constraints that limit current models, and outlines research directions to bridge artificial and biological memory systems—offering a roadmap toward architectures that not only extend computational capacity but also support genuine artificial cognition.

\subsection{Overview and Synthesis}
\textbf{Evolutionary trajectory and convergence}
\begin{itemize}
    \item Foundation (2019–2021): Early systems established explicit memory management beyond standard attention via state-based recurrence and compression, demonstrating that long-range modeling benefits from persistent activations and hierarchical reduction (e.g., Transformer-XL \cite{dai2019transformer}; Compressive Transformer \cite{rae2019compressive}).
    \item Expansion (2022–2024): Retrieval-augmented modeling scaled access from thousands to billions of entries using kNN/MIPS indexing and chunked cross-attention (e.g., Memorizing Transformer \cite{wu2022memorizing}; RETRO \cite{borgeaud2022improving}; EMAT \cite{wu2022efficient}; Memory Layers at Scale \cite{berges2024memory}), while architectures diversified to associative, hierarchical, and graph-based organization (e.g., AiT \cite{sun2023associative}, MemGPT \cite{packer2023memgpt}, MemWalker \cite{chen2023walking}, HippoRAG \cite{gutierrez2024hipporag}, CDMem \cite{gao2025efficient}). Surprise-gated updates emerged as a biologically motivated write policy (e.g., Titans \cite{behrouz2024titans}), complementing selective reset/decay and LRU strategies for stability under growth.
    \item Maturation (2025): Production-oriented designs emphasized hybrid storage, test-time adaptation, and specialized access, including expert routing and compression-based interfaces (e.g., zip2zip \cite{geng2025zip2zip}), operational memory OS abstractions (e.g., MemoryOS \cite{kang2025memory}), and hierarchical controllers for reasoning (e.g., HRM \cite{wang2025hierarchical}), reflecting a shift from static pattern recognition to adaptive, experience-driven intelligence.
\end{itemize}

\textbf{Architecture: hybrid dominance.} Parameter-encoded memory offers immediate access but risks catastrophic interference when updated; state-based memory supports rapid adaptation but is capacity-limited; external stores scale but add retrieval/consistency overhead. Hybrid designs increasingly combine these modalities to balance latency, scalability, and plasticity through division of labor and policy-driven coordination.

\textbf{Memory dynamics: from rules to policies.}
Write operations progressed from static schedules to surprise-gated consolidation and learned policies, mitigating stability–plasticity trade-offs by updating on prediction errors and adapting allocation/eviction to task demands. Test-time plasticity became the default, enabling personalization and continual adaptation in deployment.

\textbf{Retrieval and forgetting: specialization matters.}
Access evolved beyond attention and pure similarity toward structure-aware methods: graph navigation for relational queries, associative retrieval for content-addressable access, and hierarchical/expert routing for specialization and efficiency. Forgetting moved from FIFO/decay to LRU, selective, cycle-based, and evolutionary strategies, showing that intelligent erasure—aligned with utility and hierarchy—is as consequential as storage and retrieval for sustained performance under growth.

\subsection{Challenges}
Despite remarkable progress toward cognitive memory architectures, our comprehensive analysis reveals fundamental challenges that continue to constrain practical deployment and theoretical understanding of memory-augmented systems.

\textbf{Scalability and Retrieval Bottlenecks.} Despite significant architectural innovations, memory-augmented systems face fundamental scalability constraints that limit practical deployment at scale. Current approaches demonstrate distinct trade-offs between computational efficiency, retrieval accuracy, and resource requirements.

Retrieval mechanisms exhibit characteristic scaling limitations. Approximate similarity search methods, while computationally efficient, suffer from accuracy degradation as memory size increases \citep{wu2022efficient}. Product-key decomposition approaches successfully reduce lookup complexity from linear to sub-linear scaling \citep{berges2024memory}, yet encounter parameter overhead that constrains expansion to billion-entry systems. Graph-based retrieval methods enable sophisticated multi-hop reasoning but face exponential complexity growth with increasing graph density and traversal depth.

Compression-based solutions present complementary challenges. While techniques like adaptive tokenization can achieve substantial sequence reduction, they introduce inference-time computational overhead that may offset retrieval gains. Hierarchical memory organization similarly requires careful balance between compression efficiency and information fidelity.

Infrastructure considerations increasingly limit deployment viability. Memory-augmented architectures impose substantial storage bandwidth requirements, introduce novel security vulnerabilities through persistent external memory, and exhibit non-linear energy scaling patterns. Distributed implementations face additional consistency and latency challenges that can negate theoretical performance advantages. Cross-modal systems compound these issues by requiring unified similarity metrics that may inadequately represent heterogeneous data types, leading to systematic retrieval degradation across modalities.

\textbf{Memory Interference and Coordination.} 
Memory systems face fundamental coordination challenges beyond scalability, particularly in multi-task scenarios. The stability-plasticity dilemma manifests distinctly across architectures: while current literature confirms surprise-gated systems excel at novelty detection, evidence for their specific struggles with gradual knowledge drift remains limited. Policy-learned approaches show strong task adaptation capabilities but face inherent overfitting risks during continual learning.

Memory interference emerges as a critical bottleneck when similar contexts trigger conflicting information retrieval. External memory systems suffer from catastrophic collisions during concurrent access, while parameter-encoded approaches experience gradient interference during continual updates. Hybrid architectures attempt to mitigate these issues through functional partitioning, yet optimal allocation strategies remain highly domain-dependent.

Forgetting policies introduce additional coordination complexities. A-MEM's \cite{xu2025mem} evolutionary approach demonstrates promise for utility-based memory curation but requires careful tuning to avoid inadvertent erasure of rare but valuable information. Systems employing LRU and selective forgetting strategies perform effectively in structured environments but struggle under non-stationary conditions where relevance patterns shift unpredictably.

\textbf{Evaluation and Standardization Gaps.}
The field suffers from fragmented evaluation methodologies that prevent systematic cross-architecture comparisons. Current benchmarks exhibit dramatic variation in memory requirements, task complexity, and evaluation metrics, making it difficult to assess fundamental trade-offs between different memory strategies. 
Critical deficiencies include inconsistent context length protocols, divergent benchmark emphases on retrieval versus reasoning capabilities, and absent robustness evaluations for adversarial scenarios, memory corruption, and distribution shift. Most significantly, existing evaluations rarely assess long-term adaptation, memory utilization efficiency, or interference mitigation—precisely the capabilities that distinguish sophisticated memory systems from basic approaches.

\subsection{Future Directions}
The evolution of memory-augmented Transformers toward truly cognitive architectures requires coordinated advances across multiple research frontiers. The following directions represent the most promising paths for achieving human-like memory capabilities while addressing current technical limitations.

\textbf{Toward Cognitive Flexibility and Lifelong Learning.}
Emerging paradigms in memory-augmented transformers focus on building systems that can dynamically store, retrieve, and update knowledge in ways that reflect the adaptability of biological memory. Neuroscientific insights from \citet{dijksterhuis2024pronouns} highlight the value of memory consolidation, revealing how concept cells in the human hippocampus reactivate when pronouns reference specific nouns, seamlessly linking new linguistic input to stored concepts-a mechanism comparable to integrating episodic memories into an LLM’s parametric memory to bypass capacity constraints and achieve lasting retention. Complementing this, the position paper by \citet{pink2025position} argues that episodic memory is a vital missing component for long-term LLM agents, proposing a framework with five essential properties to foster adaptive behavior and outlining a research roadmap to embed these capabilities. A pivotal trend in this direction is the decoupling of computation from storage, enabling models to tap into external or hybrid memory banks for real-time, current information without the need for retraining. Such architectures facilitate personalized, context-responsive outputs in evolving environments. Furthermore, innovations like test-time training and memory-driven optimization empower models to learn and adapt during deployment, bolstered by selective forgetting and zero-shot transfer mechanisms that enhance generalization. The integration of multimodal memory and collaborative networks also holds promise for deeper reasoning and shared learning among agents. To sustain these advancements, progress in hierarchical storage, memory compression, and hardware-aware design is driving scalable, energy-efficient deployment across varied platforms.

\textbf{Toward Human-Like Cognition: The Role of Memory in Intelligent Agents.}
As intelligent agents evolve toward more human-like reasoning and autonomy, the integration of sophisticated memory systems becomes a central design challenge. Unlike conventional LLMs that operate in a stateless fashion, truly interactive agents must preserve context, interpret ongoing events, and adapt their behavior across time \citep{liang2024self,yi2025score}. Drawing from cognitive science, recent agent architectures incorporate short-term memory for maintaining dialogue context, working memory for in-the-moment reasoning, and long-term memory for accumulating knowledge and past experiences \citep{li2024falcon}. Vector databases have emerged as a popular solution for implementing long-term memory, enabling fast, similarity-based retrieval of episodic and procedural knowledge \citep{hatalis2023memory}. However, realizing robust memory-driven behavior introduces significant difficulties. Agents often fail to separate memory types, leading to conflicts between episodic and semantic recall, and may repeatedly attempt failed subtasks without effective use of episodic feedback \citep{wang2024jarvis}. As memory grows over time, retrieval speed and storage cost become critical concerns, especially when managing large volumes of data. Static or manually defined metadata can limit retrieval quality, pointing to a need for agents to learn metadata attributes dynamically to support smarter decision-making \citep{sarch2023open}. Moreover, integrating long-term memory with external knowledge bases like ontologies or knowledge graphs could enhance contextual grounding and reasoning \citep{wang2024jarvis}. Addressing these issues is essential to building agents capable of flexible, adaptive cognition that mirrors the structure and function of human memory systems.

\textbf{Future Architectures and Ethical Considerations.} Test-time training, memory-driven optimization, and zero-shot transfer learning allow models to adapt during deployment, offering the promise of lifelong learning. Multimodal memory systems and collaborative agent networks open new paths for collective intelligence, deeper reasoning, and shared learning across environments.
Yet, these capabilities also introduce ethical and societal considerations. As memory-augmented Transformers are adopted in sensitive domains like healthcare, education, and personalized services, ensuring transparency, privacy, and user control over memory becomes vital. Techniques for explainable memory operations, data auditing, and bias mitigation will be critical to build trust and prevent misuse.

In summary, the future of memory-augmented Transformers lies in bridging engineering efficiency with cognitive flexibility. By combining continual learning, dynamic memory adaptation, and biologically inspired design, alongside ethical safeguards, these systems have the potential to transform AI from static pattern recognizers into adaptive, intelligent agents.

\bibliography{reference}
\bibliographystyle{tmlr}


\end{document}